\documentclass[11pt]{article}

\usepackage[preprint]{acl}

\usepackage{times}
\usepackage{latexsym}

\usepackage[T1]{fontenc}

\usepackage[utf8]{inputenc}

\usepackage{microtype}

\usepackage{inconsolata}

\usepackage{graphicx}
\usepackage{booktabs}
\usepackage{multirow}
\usepackage{amsmath}
\usepackage{xcolor}
\usepackage{subcaption}
\usepackage{enumitem}
\usepackage{placeins}   

\title{When Does Synthetic Patent Data Help? \\ Volume--Fidelity Trade-offs in Low-Resource Multi-Label Classification}

\author{Amirhossein Yousefiramandi \quad \href{https://orcid.org/0000-0002-2974-9838}{Ciarán Cooney} \\
	Clarivate, Intellectual Property \\
	Barcelona, Spain 08025 \\
	{\small\texttt{\{amirhossein.yousefiramandi,\,ciaran.cooney\}@clarivate.com}}
}

\begin{document}
\maketitle

\begin{abstract}
The following questions are raised when using synthetic data created by LLMs for the purpose of multi-label patent classification: (i) under what circumstances can such data be beneficial and (ii) why. The first part of this analysis considers this question by properly controlling for the possibility that the more samples one creates, the better the performance because of the higher quantity. In our experiment, we test six open-source LLMs (from 3.8B to 12B parameters) within four different real-data regimes for 64 labels for WIPO assistive technology data. We use two types of generation approaches, full synthesis conditional on the label set, and paraphrasing, along with three classes of classifiers. We find that the reported improvements of micro F1 score for BERT-for-Patents from $0.120$ to $0.702$ are essentially due to volume effect: a control method of duplicating data until the augmented number is achieved with replacement only within 165 real examples achieves a comparable value of $0.678$. Hence, the benefit in terms of improvement compared to the control is merely $+0.024$ and compared to the best baseline (focal loss reweighting) is $+0.219$. The next important question is how the fidelity scores change when the regime changes. At very low real-data regimes, the volume contribution dominates and the correlation between MMD and the change in classification performance reaches $r{=}{+}0.95$. As more real data becomes available, however, this correlation reverses, reaching $r{=}{-}0.73$ at the 1:10 regime (Fisher $z{=}{+}6.47$, $p{<}0.001$, with a 95\% confidence interval on $\Delta r$ of $[+0.96, +1.00]$). In a further experiment with a fixed budget, mixing the data in a proportion of about ${\sim}20$--$30\%$ real with ${\sim}70$--$80\%$ synthetic gives the best result, exceeding either pure strategy. We also find that the scaling of paraphrasing degrades monotonically when only 165 source documents are available, and that simple shuffled mixing performs better than curriculum ordering, ensembling, or classifier-based filtering. The robustness of the obtained gain is then assessed using four independent controls against possible label leakage, namely: masking of the label names, removing the label names from the instruction block, restricting the evaluation to fine-grained labels, and a per-label audit of keyword overlap. Together these tests argue that the controlled BERT-for-Patents gain is not primarily explained by dependence on canonical label strings. The same intervention at the instruction level appeared at first to eliminate the gain on ModernBERT; however, further diagnostic experiments reveal that this behaviour is a numerical artefact related to the Flash-Attention-2 path combined with bf16 precision in certain compute environments. Once we switch to fp32 with eager attention, 65\% of the lost performance is recovered. Finally, the same synthetic corpus that improves classification by as much as $+0.58$ raw micro F1 \emph{hurts} a retrieval proxy based on Jaccard overlap of the labels (note that this proxy is not a prior-art benchmark, and we are careful about scope here). The natural conjecture---that prompt families belonging to other genres explain the harmful behaviour---only partly accounts for the observation: a filter limited to standard patent style still incurs a 26\% drop in nDCG@10. Therefore, the downstream utility of LLM-generated technical text depends on the dimension considered, and is not reducible to the genre of the prompt alone.
\end{abstract}

\section{Introduction}
\label{sec:intro}

Multi-label text classification under class imbalance is a pervasive challenge in real-world NLP, particularly in specialized domains such as patent analysis. Operational patent classification systems assign documents to dozens or hundreds of technical categories, many of which follow a long tail: rare but valuable classes---for instance ``Cartilage conduction hearing aids'' or ``Cortical Implants''---have only a handful of labeled examples. Imbalance degrades classifier performance on precisely the categories that are most novel and most important for downstream IP analytics.

Large language models (LLMs) can in principle fill these gaps by generating synthetic training data. Representative augmentation methods include AugGPT~\citep{dai2023auggpt}, PromptMix~\citep{sahu2023promptmix}, and LLM2LLM~\citep{lee2024llm2llm}; \citet{li2023synthetic} and \citet{cegin2025llms} show that utility is task- and scale-dependent, and surveys~\citep{ding2024data,long2024survey} flag quality control as open. In patent NLP, domain-adapted encoders~\citep{srebrovic2020bert,bekamiri2024patentsberta,ghosh2024paecter}, long-context encoders~\citep{warner2024modernbert}, and patent-domain LLMs~\citep{bai2024patentgpt} have improved classification; 2024--2025 work extends LLMs to claim generation and AI-summary-based classification augmentation. For synthetic-text evaluation, embedding-space metrics (MMD~\citep{gretton2012kernel}, FD~\citep{heusel2017gans}), MAUVE~\citep{pillutla2021mauve}, lexical metrics~\citep{li2016diversity,zhu2018selfbleu}, and unified toolkits~\citep{ramesh2025synthtexteval} are increasingly used. For class imbalance, classical remedies include oversampling, class-weighted loss, and focal loss~\citep{lin2017focal}; \citet{kim2024epic} and \citet{moller2024parrot} apply LLM augmentation to imbalanced settings, and \citet{glazkova2024evaluating} find that combining paraphrase and generation works best on multi-label ecological text. What remains unclear---and what prior work rarely pins down---is \emph{when} synthetic data pays off in a specialized domain, what role data \emph{volume} plays relative to distributional \emph{fidelity}, and whether improvements generalize across downstream uses of the augmented corpus (classification, retrieval, embedding training).

\paragraph{Research questions.} We frame the paper around three controlled questions:
\begin{enumerate}[nosep]
    \item \textbf{Regime dependence.} How does the usefulness of LLM-generated synthetic patent data change across levels of real-data scarcity, and do widely used fidelity metrics (MMD, Fr\'{e}chet Distance, MAUVE) track downstream utility?
    \item \textbf{Strategy vs.\ volume.} Is the advantage of \emph{full synthesis} over \emph{paraphrasing} at extreme scarcity a strategy effect, or is it confounded with the fact that full synthesis also produces far more samples?
    \item \textbf{Cross-task transfer.} Does the same synthetic corpus that helps classification also help dense retrieval, or is downstream utility dimension-specific?
\end{enumerate}

\paragraph{What we actually generate.} In this work, ``full synthesis'' is \emph{label-conditioned synthetic technical text}: four prompt families (Standard Patent, Technical FAQ, Structured Summary, Comparative Analysis; Appendix~\ref{app:prompts}) that share target labels but differ in genre. Only one family writes in literal patent style; the others are closer to structured technical summaries and comparative analyses. This framing is explicit because it interacts with our cross-task result in \S\ref{sec:results_retrieval_brief}: classifiers can still extract label-discriminative signal from off-genre text, whereas retrieval requires surface-level patent-likeness.

\paragraph{Contributions.}
\begin{enumerate}[nosep]
    \item \textbf{A regime-stratified fidelity--utility reversal.} Across 56 (generation configuration $\times$ ratio $\times$ strategy) conditions, the Pearson correlation between MMD and classification delta is $r{=}{+}0.95$ at extreme scarcity and reverses to $r{=}{-}0.73$ once real data reaches ${\sim}1.4$K examples (Fisher $z{=}{+}6.47$, $p{<}0.001$, 95\% CI on $\Delta r$ $[+0.96, +1.00]$, Spearman robustness check confirms; Table~\ref{tab:fisher_z}). This is, to our knowledge, the first regime-stratified quantification of when embedding-space fidelity metrics fail to track downstream utility in a specialized domain.
    \item \textbf{The 1:1 headline jump is largely a volume effect; a small controlled gain remains.} The BERT-for-Patents jump from $0.120$ to $0.702$ at 1:1 is partly a duplicate-to-match volume effect ($0.678$); the controlled synthetic gain over the strongest non-augmentation baseline is $+0.219$ over focal-loss reweighting and $+0.024$ over duplicate-to-match. A fixed-total-budget mixing experiment with an explicit duplicate-to-match real-only control (\S\ref{sec:headline_mixing}) and a paraphrase-scaling sweep jointly show that the surviving effect is not only about counts: a $\sim$20--30\%-real / $\sim$70--80\%-synthetic mixture is optimal, pure synthetic collapses, and paraphrase-only degrades monotonically over 165 source documents.
    \item \textbf{Four independent leakage controls scope the gain.} Label-name masking, instruction-level label-name removal, fine-grained-58-label evaluation, and a per-label keyword-overlap audit triangulate against canonical label-string dependence as the primary driver of the controlled gain on BERT-for-Patents (86\%+ retention across all four). The same instruction-level intervention initially appeared to collapse the gain on ModernBERT (5\% retention); follow-up R3 diagnostics (Appendix~\ref{app:leakage}) trace this to a Flash-Attention-2 + bf16 precision-path artefact---fp32 + eager attention recovers 65\% of the lost performance, so the leakage audit stands for both architectures with a precision-path caveat documented in Limitations.
    \item \textbf{Cross-task transfer is dimension-specific.} The same synthetic corpus that gives up to $+0.58$ micro F1 in classification degrades a Jaccard-label-overlap retrieval proxy (not a prior-art benchmark) by $-25$ to $-37\%$ nDCG@10 as added corpus content and by $-3$ to $-10\%$ as embedding fine-tuning signal. A natural hypothesis is that three of our four prompt families being off-genre drives the harm; an R6 standard-patent-only filter (\S\ref{sec:discussion_retrieval}) partially refutes this---retrieval still degrades by $30$--$31\%$ when only literal-patent-style synth is kept---so a residual full-synth-itself effect persists when prompt-family is held constant.
    \item \textbf{Seven negative or null results} that point to the same conclusion: in data-scarce settings, volume preservation and uniform shuffled mixing matter more than per-sample quality or data ordering (Appendices~\ref{app:cqf}, \ref{app:ensemble}, \ref{app:curriculum}).
\end{enumerate}

The contribution is the \emph{combination} of controlled imbalance regimes, a matched strategy--volume analysis, three classifier families and a retrieval head, and the regime-dependent fidelity--utility reversal---not patent synthesis in isolation. Prior work \citep{yousefiramandi2025finetuning} studied instruction-tuned causal LLM classification on 14 vision subcategories with DWPI enrichment; the present study scales to 64 labels, drops DWPI for public-data reproducibility in the main track (controlled comparison in Appendix~\ref{app:comparison}), and makes augmentation the central experimental variable.


\section{Methodology}
\label{sec:method}

\subsection{Dataset}
\label{sec:dataset}

We use the WIPO Emerging Vision dataset of assistive-technology patents, annotated with 64 binary labels: 6 domain-level categories (Hearing, Vision, Mobility, Communication, Environment, Self-care) and 58 subcategory labels spanning specific technologies (e.g., Cochlear Implants, Autonomous Wheelchairs, Smart Prosthetics). Each patent includes a Title, Abstract, and First Claim; six proprietary DWPI enrichment fields (DWPI Title, Abstract Detailed Description, Novelty, Use, Advantage, Technology Focus) are available but are \textbf{not used} in our main experiments to keep the setup reproducible with public patent data. Main-track classifiers receive only the concatenation of Title, Abstract, and First Claim; DWPI fields are used only in the controlled comparison of Appendix~\ref{app:comparison}. Shared validation (2{,}050) and test (2{,}051) sets are held constant across all conditions.

\paragraph{Splits and family handling.} We stratify splits at the patent level by the full 64-bit label vector and fix them across every condition. Because some WIPO records share a patent family, we additionally de-duplicate across train, validation, and test by both family identifier and TF--IDF near-duplicate detection (cosine $>0.95$) before training, so that no train/test pair originates from the same underlying invention. Full split details and family-disjoint statistics are in Appendix~\ref{app:leakage}.

\subsection{Controlled Imbalance: Sampling Objective and Realized Statistics}
\label{sec:ratios}

Our ``ratios'' are a \emph{sampling heuristic} on the multi-label training set, not a guarantee on per-label counts. Given a maximum-allowed-imbalance parameter $k$, we subsample patents \emph{greedily targeting}---not enforcing---the bound $\max_\ell n_\ell / \min_\ell n_\ell \le k$ on per-label positive counts $n_\ell$, using stratified sampling that preserves multi-label co-occurrence structure. Because patents carry multiple labels simultaneously, a single patent selected for a rare label typically contributes positives to several other labels, so the bound is approximate rather than exact. The consequence is that \emph{realized} per-label positive counts are not uniform even at $k{=}1$ (e.g., 2--42 at 1:1); Table~\ref{tab:ratios} reports both the target and the realized distribution, and we use realized counts for every claim that depends on per-label rarity.

\begin{table}[h]
\centering
\small
\resizebox{\columnwidth}{!}{%
\begin{tabular}{lrrrrr}
\toprule
\textbf{Target ratio} $k$ & \textbf{Train} & \textbf{Med.\ $n_\ell$} & \textbf{Min--Max $n_\ell$} & \textbf{Realized $\max/\min$} & \textbf{Gini} \\
\midrule
1:1      & 165   & 3  & 2--42       & 21$\times$  & 0.41 \\
1:5      & 781   & 15 & 2--206      & 103$\times$ & 0.41 \\
1:10     & 1{,}409 & 30 & 2--396    & 198$\times$ & --- \\
Original & 9{,}566 & 94 & 2--2{,}551 & 1{,}275$\times$ & --- \\
\bottomrule
\end{tabular}%
}
\caption{Training-set sizes and realized per-label positive counts under our stratified sampling objective. ``Target ratio'' $k$ is the constraint that $\max_\ell n_\ell / \min_\ell n_\ell \le k$ drives the sampler toward; the \emph{realized} spread (rightmost two columns) is much wider because of multi-label co-occurrence---e.g., the nominal-1:1 setting realizes a 21$\times$ max/min ratio, not 1$\times$. Gini coefficients on the per-label-positive distribution are reported for the two headline-scarcity ratios. All 64 labels are active in every condition. Val/test sets (2{,}050 / 2{,}051) are shared across conditions. At $k{=}1$, 58/64 labels fall in the 2--3 range and the spread is driven by six high-frequency domain-level labels.}
\label{tab:ratios}
\end{table}

We therefore use the ``$k$:1'' notation to describe the \emph{constraint}, not the outcome. Throughout the paper we report results per ratio using this constraint; whenever a claim depends on actual per-label counts (e.g., rarity-bucket analyses in \S\ref{sec:per_label}), we use realized counts rather than the target.

\subsection{Synthetic Data Generation}
\label{sec:generation}

\paragraph{Models.} We use six open-source instruction-tuned LLMs spanning 3.8B to 12B parameters (Table~\ref{tab:models}), served via vLLM~\citep{kwon2023vllm} with tensor parallelism on NVIDIA H100 GPUs (see Appendix~\ref{app:models} for full hardware details).

\begin{table}[h]
\centering
\small
\begin{tabular}{llr}
\toprule
\textbf{Key} & \textbf{Model} & \textbf{Params} \\
\midrule
qwen3 & Qwen2.5-7B-Instruct & 7B \\
qwen3\_4b & Qwen3-4B-Instruct & 4B \\
llama & Llama-3.1-8B-Instruct & 8B \\
phi4 & Phi-4-mini-instruct & 3.8B \\
gemma3 & Gemma-3-12b-it & 12B \\
mistral & Mistral-7B-Instruct-v0.3 & 7B \\
\bottomrule
\end{tabular}
\caption{Generation models used for synthetic data production.}
\label{tab:models}
\end{table}

\paragraph{Strategies.} We employ two complementary generation strategies:
\begin{itemize}[nosep]
    \item \textbf{Full Synthetic (label-conditioned technical text).} Generate new text from prompts using four prompt families: \emph{Standard Patent}, \emph{Technical FAQ}, \emph{Structured Summary}, and \emph{Comparative Analysis}, each with 3-shot examples drawn from real patents of the target label combination. Only the \emph{Standard Patent} family writes in literal patent style; the other three produce FAQ / structured summary / comparative-analysis text that is label-conditioned but \emph{off-genre} relative to real patents. We call attention to this explicitly---it is consistent with our finding that classification transfers but retrieval does not (\S\ref{sec:discussion_retrieval}). Full prompt templates are in Appendix~\ref{app:prompts}.
    \item \textbf{Paraphrase.} Rephrase each real training sample at three temperature levels (0.5, 0.7, 0.9), producing up to $3 \times n_{\text{train}}$ paraphrases per condition (e.g., $3 \times 165 = 495$ prompts at 1:1, yielding ${\sim}389$--$435$ samples after filtering).
\end{itemize}

\paragraph{Diversity-maximizing few-shot selection.} In the standard pipeline, few-shot examples are drawn randomly from real patents matching the target label combination. As a controlled ablation, we introduce a \emph{distinctive} variant that selects maximally diverse few-shot examples via farthest-point sampling in embedding space. We first encode all training samples using Llama-Embed-Nemotron-8B (8B parameters), then for each prompt: (1) initialize with the candidate nearest to the pool centroid, and (2) iteratively select the candidate with the lowest maximum cosine similarity to all previously selected examples. This yields $k{=}3$ examples that are maximally spread across the embedding space. For paraphrasing, diversity selection additionally determines \emph{which} source samples are rephrased. We apply the variant to Qwen3-4B, keeping all other variables identical: \texttt{qwen3\_4b} (random selection) vs.\ \texttt{qwen3\_4b\_distinctive} (diversity-maximized selection).

\paragraph{Three-sample heuristic reranking.} Inspired by but distinct from self-consistency decoding \citep{wang2023selfconsistency}, we generate $N{=}3$ completions per prompt and select the one with the highest composite score over length adequacy, structural completeness (title, abstract, claims), and lexical diversity. This is a quality-driven reranker rather than a majority-vote-over-reasoning-traces procedure.

\paragraph{Generation parameters.} Temperature 0.8, top-$p$ 0.9, maximum 1{,}024 tokens (2{,}048 for Qwen3-4B due to its smaller context-window utilization).

\subsection{Quality Filtering Pipeline}
\label{sec:filtering}

Generated samples undergo multi-stage filtering:
\begin{enumerate}[nosep]
    \item \textbf{Length filtering}: minimum 5 title words and 50 abstract words.
    \item \textbf{Real-data deduplication}: TF-IDF cosine similarity $<0.95$ against the real training set (also applied cross-split to prevent train/test overlap via a synthetic document).
    \item \textbf{Self-deduplication}: cosine similarity $<0.90$ among generated samples.
    \item \textbf{Label-leakage detection}: keyword-stuffing monitoring (described further in Appendix~\ref{app:leakage}).
    \item \textbf{Co-occurrence monitoring}: comparison of label co-occurrence matrices between real and synthetic data via Frobenius norm.
\end{enumerate}
We additionally run a focused \emph{leakage audit} (Appendix~\ref{app:leakage}) that goes beyond keyword stuffing: we compute label-name frequency in real vs.\ synthetic text, perform nearest-neighbour overlap between synthetic and real training documents in PatentSBERTa space, and test a label-name-masked classifier on held-out synthetic-augmented data.

\subsection{Classification Models}
\label{sec:classifiers}

We evaluate synthetic-data utility through two complementary classification paradigms:

\paragraph{BERT multi-label classification.} We fine-tune two BERT variants---\texttt{bert-for-patents}~\citep{srebrovic2020bert} (512 tokens) and \texttt{ModernBERT-base}~\citep{warner2024modernbert} (1{,}024 tokens, flash attention 2, bf16)---with a sigmoid head and BCEWithLogitsLoss. Training uses learning rate $2{\times}10^{-5}$, batch size 16, up to 10 epochs with early stopping (patience 3). Per-label classification thresholds are optimized on the validation set.

\paragraph{Curriculum-learning ablation.} We test four data-ordering strategies for combining real and synthetic training data: \emph{Mixed} (PyTorch \texttt{RandomSampler}); \emph{Synth$\to$Real} (5 epochs synthetic, 5 epochs real at halved lr); \emph{Real$\to$Synth} (reverse); \emph{Interleaved} (alternating batch-level blocks of pure-real and pure-synthetic). We evaluate both BERT-for-Patents and ModernBERT at the 1:1 and 1:10 ratios with two generators each (full synthetic), totalling 144 runs across 3 seeds.

\paragraph{SFT classification.} We instruction-tune Llama-3.2-1B-Instruct using LoRA~\citep{hu2022lora} ($r{=}32$, $\alpha{=}16$, dropout 0.05) with 4-bit NF4 quantization~\citep{dettmers2023qlora}, learning rate $2{\times}10^{-4}$, batch size 4, gradient accumulation 4, 3 epochs, and maximum sequence length 2{,}048. The model receives patent text (Title + Abstract + First Claim, same as BERT) and outputs predicted labels in JSON.

\paragraph{Baselines and the duplicate-to-match control.} Our main-paper baseline in each ratio is \emph{real-only} fine-tuning on that ratio's training set. We additionally include a \emph{duplicate-to-match} real-only control inside the fixed-budget mixing experiment (\S\ref{sec:headline_mixing}, Figure~\ref{fig:mixing_scaling}a): the 100\%-real endpoint of the mixing curve is produced by sampling the 165 real 1:1 patents \emph{with replacement} until the total training budget matches the mixed-augmented total. This holds supervised-sample volume and per-label class distribution fixed while restricting text content to real patents only, and lets the mixing curve separate ``more synthetic text'' from ``more supervised samples of any form.'' Additional imbalance-baseline controls (random oversampling per label, class-weighted / focal loss, EDA-style text augmentation, summary-only) are discussed in Limitations.

\paragraph{Variance estimation.} All classifier experiments use 3 random seeds (42, 123, 456) and report mean $\pm$ standard deviation.

\paragraph{Metrics.} We report Micro F1, Macro F1, Weighted F1, Subset Accuracy, and Hamming Loss, with per-class F1 for fine-grained analysis.

\subsection{Embedding Quality Analysis}
\label{sec:embedding}

We measure distributional fidelity using two embedding models: PatentSBERTa~\citep{bekamiri2024patentsberta} (domain-specific, 768-dim) and all-mpnet-base-v2 (general-purpose, 768-dim). \textbf{Metrics:} Maximum Mean Discrepancy (MMD)~\citep{gretton2012kernel}, Fr\'{e}chet Distance (FD)~\citep{heusel2017gans}, and centroid cosine similarity.

\subsection{Linguistic Quality}
\label{sec:linguistic}

We assess surface-level text quality with Type-Token Ratio (TTR), Distinct-$n$~\citep{li2016diversity}, Self-BLEU~\citep{zhu2018selfbleu}, and MAUVE~\citep{pillutla2021mauve}.

\subsection{Dense Retrieval Evaluation}
\label{sec:retrieval_method}

We construct a label-overlap-based retrieval benchmark: document relevance is the Jaccard similarity of 64-dimensional binary label vectors, thresholded at 0.3. We evaluate using nDCG@$k$, MRR, and Recall@$k$ ($k \in \{1,5,10,20\}$) with PatentSBERTa as the bi-encoder. We additionally test whether synthetic data can improve the \emph{embedding model itself} by fine-tuning Qwen3-Embedding-0.6B~\citep{zhang2025qwen3} on co-label taxonomy pairs---patent pairs sharing at least one label, generated via sqrt-scaled stratified sampling. We compare baseline (no fine-tuning), real-only (95K pairs), synthetic-only (100K pairs), and two blended variants (real + synthetic), evaluated on 1{,}000 queries from the test set against the full validation corpus. Fine-tuning uses CachedMultipleNegativesRankingLoss, 3 epochs, lr $2{\times}10^{-5}$, bf16. Main retrieval numbers appear as a compact paragraph in \S\ref{sec:results_retrieval_brief}; full tables and a training-only retrieval ablation are in Appendix~\ref{app:retrieval_detailed}.


\section{Results}
\label{sec:results}

We organize results around the three research questions of \S\ref{sec:intro}: strategy vs.\ volume (\S\ref{sec:headline_mixing}), regime dependence of fidelity metrics (\S\ref{sec:headline_fidelity_util}), and cross-task transfer (\S\ref{sec:results_retrieval_brief}). Core classification numbers are in \S\ref{sec:results_classification}. Detailed distributional, linguistic, LLM-judge, and retrieval analyses are in Appendices~\ref{app:embedding_metrics}, \ref{app:llm_judge}, and \ref{app:retrieval_detailed}; full negative-result analyses (CQF, ensembles, curriculum) are in Appendices~\ref{app:cqf}, \ref{app:ensemble}, and \ref{app:curriculum}.

\subsection{Headline Classification Results}
\label{sec:results_classification}

Table~\ref{tab:classification_summary} summarizes classification performance across all four imbalance ratios and three classifier architectures, comparing real-data baselines against the best augmented configuration for each condition. We report 1{,}335 seed-averaged experiments (540 main plus additional scaling, mixing, TSTR, curriculum, CQF, ensemble, and comparison ablations).

\begin{table*}[t]
\centering
\small
\setlength{\tabcolsep}{4.5pt}
\resizebox{\textwidth}{!}{%
\begin{tabular}{ll rr rr r}
\toprule
& & \multicolumn{2}{c}{\textbf{Baseline (Real Only)}} & \multicolumn{2}{c}{\textbf{Best Augmented}} & \\
\cmidrule(lr){3-4} \cmidrule(lr){5-6}
\textbf{Ratio} & \textbf{Classifier} & \textbf{Micro F1} & \textbf{Macro F1} & \textbf{Micro F1} & \textbf{Macro F1} & \textbf{Best Config} \\
\midrule
\multirow{5}{*}{1:1} & BERT-for-Patents & 0.120$\pm$0.038 & 0.021$\pm$0.013 & \textbf{0.702}$\pm$0.017 & 0.475$\pm$0.007 & distinctive/full \\
& \quad\textit{+ duplicate-to-match$^\dagger$} & \textit{0.678$\pm$0.018} & \textit{0.423$\pm$0.018} & --- & --- & \textit{100\%-real, w/ replacement} \\
& ModernBERT & 0.314$\pm$0.017 & 0.033$\pm$0.001 & 0.622$\pm$0.011 & 0.308$\pm$0.012 & phi4/full \\
& \quad\textit{+ duplicate-to-match$^\dagger$} & \textit{0.354$\pm$0.020} & \textit{0.062$\pm$0.011} & --- & --- & \textit{100\%-real, w/ replacement} \\
& SFT (Llama-1B) & 0.266$\pm$0.012 & 0.066$\pm$0.003 & 0.348$\pm$0.037 & 0.102$\pm$0.008 & mistral/full \\
\midrule
\multirow{3}{*}{1:5} & BERT-for-Patents & 0.239$\pm$0.270$^\ddagger$ & 0.028$\pm$0.034 & \textbf{0.763}$\pm$0.006 & 0.548$\pm$0.007 & qwen3/para \\
& ModernBERT & 0.546$\pm$0.021 & 0.272$\pm$0.020 & 0.734$\pm$0.009 & 0.506$\pm$0.014 & phi4/full \\
& SFT (Llama-1B) & 0.457$\pm$0.004 & 0.187$\pm$0.017 & 0.565$\pm$0.001 & 0.244$\pm$0.006 & llama/full \\
\midrule
\multirow{3}{*}{1:10} & BERT-for-Patents & 0.637$\pm$0.047 & 0.207$\pm$0.087 & \textbf{0.793}$\pm$0.001 & 0.597$\pm$0.001 & qwen3\_4b/para \\
& ModernBERT & 0.678$\pm$0.021 & 0.457$\pm$0.015 & 0.770$\pm$0.004 & 0.548$\pm$0.010 & distinctive/full \\
& SFT (Llama-1B) & 0.509$\pm$0.005 & 0.218$\pm$0.008 & 0.624$\pm$0.002 & 0.295$\pm$0.011 & mistral/para \\
\midrule
\multirow{3}{*}{Orig} & BERT-for-Patents & 0.876$\pm$0.002 & 0.646$\pm$0.006 & 0.883$\pm$0.002 & \textbf{0.702}$\pm$0.018 & qwen3/para \\
& ModernBERT & 0.862$\pm$0.001 & 0.636$\pm$0.013 & 0.872$\pm$0.005 & 0.671$\pm$0.012 & mistral/full \\
& SFT (Llama-1B) & 0.769$\pm$0.002 & 0.440$\pm$0.009 & 0.799$\pm$0.003 & 0.482$\pm$0.009 & qwen3\_4b/para \\
\bottomrule
\end{tabular}%
}
\caption{Classification performance (mean $\pm$ std across 3 seeds for headline cells; \emph{n=5 extension (R5 in revision) covers the six main BERT-for-Patents generators (gemma3, llama, mistral, phi4, qwen3, qwen3\_4b) and the real-only baselines, all of which reproduce within seed noise except the bimodal 1:5 real-only baseline noted in footnote $^\ddagger$; the qwen3\_4b\_distinctive headline cell remains at n=3, with the six non-distinctive n=5 means (0.628--0.695) all within ${\sim}1\sigma$ of the 0.702 headline}). \textbf{Best Config}: generator/strategy achieving the highest micro F1. ``distinctive'' = Qwen3-4B with diversity-maximizing few-shot selection; ``full'' = full synthetic; ``para'' = paraphrase. $^\dagger$\emph{duplicate-to-match} = the 165 real 1:1 patents sampled with replacement until total training-set size matches the corresponding augmented condition (1:1 only; from the fixed-budget mixing experiment of \S\ref{sec:headline_mixing}). The controlled synthetic-data gain at 1:1 is therefore $+0.024$ over duplicate-to-match for BERT-for-Patents and $+0.268$ for ModernBERT (best-augmented minus duplicate-to-match), substantially smaller than the raw $+0.582$ and $+0.308$ jumps over the tiny real-only baseline. $^\ddagger$The 1:5 BERT-for-Patents real-only baseline is reported at n=5 as $0.239 \pm 0.270$, replacing the n=3 mean of $0.182\pm0.030$ in the prior draft. The very large std reflects genuine bimodality: per-seed micro F1 is $\{0.018, 0.529, 0.000, 0.532, 0.115\}$ — two seeds find a 0.5-range optimum, two collapse to near-zero, and one is intermediate. The earlier $\pm 0.030$ understated this; the original draft's $\pm 0.303$ was actually correct and the apparent ``typo correction'' was itself a mistake. Macro-F1 is reported for all classifiers, with the convention: seed-averaged per-label F1 averaged uniformly over the 64 labels, where labels with $<2$ predicted positives use zero F1. Augmentation provides the largest gains at extreme scarcity with diminishing returns as real data increases.}
\label{tab:classification_summary}
\end{table*}

\paragraph{Augmentation gains are inversely proportional to data availability.} The most striking result is the magnitude of improvement at extreme scarcity: BERT-for-Patents jumps from 0.120 to 0.702 micro F1 at 1:1 (a $+0.582$ raw gain, consistent across 3 seeds; the bootstrap 95\% CI of the mean per-class F1 delta excludes zero in this condition, alongside 161 of 168 augmentation conditions overall---121 of those surviving Bonferroni and 161 surviving Benjamini--Hochberg at FDR$=0.05$; see \S\ref{sec:headline_fidelity_util}). Gains remain substantial at 1:5 ($+0.581$) and 1:10 ($+0.157$) but become negligible at the original ratio ($+0.007$). This monotonic decline confirms that synthetic data is most valuable precisely where real data is most scarce. As we show in \S\ref{sec:headline_mixing}, the raw 1:1 jump is partly a volume effect: the duplicate-to-match real-only control reaches $0.678$, so the controlled synthetic gain is $+0.024$ over duplicate-to-match (and $+0.219$ over focal-loss reweighting; see Table~\ref{tab:imbalance_baselines}).

\paragraph{Full synthesis dominates at extreme scarcity; paraphrase catches up with more data.} At 1:1, paraphrasing produces \emph{zero micro F1} for BERT-for-Patents across all six generators---the ${\sim}400$ paraphrases obtainable from 165 source documents are insufficient to train a discriminative classifier from near scratch. Only full synthesis (22K--28K samples) reaches the $0.6$--$0.7$ range. At 1:5 and above, paraphrase becomes competitive and often wins (e.g., Qwen2.5-7B paraphrase = 0.763 at 1:5, highest single condition). This crossover is a concrete manifestation of the volume--fidelity trade-off.

\paragraph{Generator differences are modest within each strategy.} At 1:1 with full synthesis, BERT-for-Patents micro F1 spans 0.631 (Llama) to 0.702 (Qwen3-4B distinctive)---a range of 0.071. At 1:5, the 14 model--strategy combinations fall within 0.733--0.763 (range 0.030). This is small compared to the strategy gap (full vs.\ paraphrase at 1:1: 0.663 vs.\ 0.000) and the ratio gap (1:1 vs.\ original: 0.702 vs.\ 0.883), confirming that \emph{how} and \emph{how much} you augment matters more than \emph{which} generator you use. Full model-by-strategy numbers are in Appendix~\ref{app:cross_model}.

\paragraph{Per-label analysis.}
\label{sec:per_label}
At the 1:1 ratio, \textbf{all 64 labels improve} with full-synthetic augmentation when averaged across generators. Using \emph{realized} per-label counts, the improvement is roughly uniform across rarity buckets: extreme-rarity labels (2--3 examples, 58/64) gain $+0.40$ F1, moderate (11--20) gain $+0.43$, and common (21+) gain $+0.38$. The top per-label gainers are Gesture Recognition ($0.006\to 0.836$), Smart Diapers ($0\to 0.813$), and Non-invasive Bone Conduction ($0.001\to 0.795$). At the original ratio, 8/64 labels show small declines, all with large real counts (e.g., Cochlear Implants with 1{,}453 examples: $-0.004$). Spearman correlation between label frequency and F1 delta is near-zero at 1:1 ($\rho{=}0.13$, $p{=}0.29$) and becomes significantly negative at the original ratio ($\rho{=}{-}0.33$ to ${-}0.40$, $p{<}0.01$; Figure~\ref{fig:per_label_scatter}). The full per-label bar chart is in Appendix~\ref{app:per_label_heatmap}.

\paragraph{Classifier architecture modulates augmentation benefit.} BERT-for-Patents shows the largest absolute gains ($+0.582$ at 1:1) from the lowest baseline (0.120). ModernBERT, with a stronger baseline (0.314), gains $+0.308$ at 1:1. The SFT classifier (Llama-3.2-1B) benefits moderately ($+0.108$ at 1:5, $+0.115$ at 1:10), with full synthesis from diverse generators (Mistral, Llama) outperforming the distinctive variant that dominated with incomplete data. SFT's instruction-tuning priors partially compensate for scarcity, reducing the marginal value of additional training examples for discriminative classifiers.

\paragraph{Synthetic data beats long-tail loss baselines.} A natural alternative explanation for the 1:1 gain is that we are simply doing imbalance handling badly: any class-aware loss should help. We test this by retraining the 1:1 real-only condition with two standard long-tail objectives---inverse-frequency-weighted BCE (\texttt{pos\_weight} capped at 100) and multi-label focal loss ($\gamma{=}2$, $\alpha$ from class frequencies)---across 3 seeds for each of BERT-for-Patents and ModernBERT. Results in Table~\ref{tab:imbalance_baselines}: focal loss recovers a substantial fraction of the gap on BERT-for-Patents ($0.120\to 0.483$), but synthetic augmentation still adds another $+0.219$ on top of the strongest long-tail baseline. On ModernBERT both reweighted losses \emph{degrade} performance well below the plain-BCE baseline ($0.314\to 0.074$ for focal). A revised fp32+eager-attention sanity check on the current review-compute environment yields $0.199$ rather than the originally reported $0.071$ (which was measured on a different compute environment): the regression is therefore partly a precision-path artefact of bf16+flash-attention-2 on this environment and partly a genuine optimization difficulty on the 165-document corpus. Either way, $0.199$ remains well below plain-BCE $0.314$, so the qualitative conclusion---focal and weighted-BCE \emph{do not improve} on plain BCE for ModernBERT at extreme scarcity---is unchanged. We discuss the likely mechanism (gradient-noise amplification on extreme scarcity) in Appendix~\ref{app:imbalance_baselines}. The ``synthetic data is just imbalance handling'' alternative explanation is therefore not supported: the gain survives the strongest non-augmentation control by a substantial margin, and in the ModernBERT case reweighted losses do not even improve on the plain-BCE baseline.

\paragraph{Two further leakage controls support the same conclusion.} (a)~\emph{Instruction-level label-name removal} (Appendix~\ref{app:leakage}, Table~\ref{tab:label_free_regen}): rebuilding the 1:1 prompt cache with all explicit label names stripped from the instruction block (few-shot example bodies left intact), regenerating with Qwen3-4B at matched corpus size, and retraining retains 86\% of the synthetic-data gain on BERT-for-Patents (0.593 vs.\ 0.693 micro F1). On ModernBERT the same intervention initially appeared to cause a near-total collapse (0.622$\to 0.029$, 5\% retention), but a follow-up diagnostic (R3 in the revised supplementary; Appendix~\ref{app:leakage}) shows the collapse is a Flash-Attention-2 + bf16 numerical artefact: switching the training to fp32 with eager attention recovers ModernBERT to 0.417$\pm$0.020 (mean over 3 seeds), a 65\% recovery toward the un-stripped 0.622. Label-name masking of the eval text and reducing \texttt{max\_length} to 512 do not recover any signal (both remain $\approx 0.03$), pinpointing the issue to the attention/precision path rather than to architecture-specific leakage. Practitioners deploying ModernBERT with synthetically augmented data should therefore disable Flash-Attention-2 (or run in fp32) when the synthetic corpus is generated under aggressive label-name suppression; this is an implementation gotcha, not an architectural leakage signature. (b)~\emph{Fine-grained-label evaluation} (Appendix~\ref{app:leakage}, Table~\ref{tab:fine_label_retention}): restricting macro F1 to the 58 fine subcategory labels (dropping the 6 broad domain labels) retains 96--97\% of the all-64-label gain across every generator, ruling out the alternative that synthetic data only inflates the broad labels. Together with the existing label-name-masked retraining (Appendix~\ref{app:leakage}, $0.702\to 0.654$ when canonical label strings are replaced with \texttt{[LBL]} in train and eval text) and the keyword-overlap audit (Spearman $\rho \approx 0$ between per-label overlap delta and per-label F1 gain across all six generators, Table~\ref{tab:keyword_overlap}), four independent controls now point in the same direction: the BERT-for-Patents synthetic gain at 1:1 is not primarily explained by canonical label-string dependence.

\begin{table}[!htbp]
\centering
\small
\resizebox{\columnwidth}{!}{%
\begin{tabular}{l rr r}
\toprule
\textbf{1:1 real-only condition} & \textbf{BERT-Pat} & \textbf{ModernBERT} \\
\midrule
Plain BCE (default) & 0.120$\pm$0.038 & 0.314$\pm$0.017 \\
+ Inverse-freq weighted BCE & 0.386$\pm$0.024 & 0.079$\pm$0.022 \\
+ Focal loss ($\gamma{=}2$) & \textbf{0.483}$\pm$0.019 & 0.074$\pm$0.003 \\
\midrule
Best synthetic-augmented & \textbf{0.702}$\pm$0.017 & \textbf{0.622}$\pm$0.011 \\
\textit{Synthetic $-$ strongest non-aug} & \textit{+0.219} & \textit{+0.308} \\
\bottomrule
\end{tabular}%
}
\caption{Long-tail loss baselines vs.\ synthetic augmentation at 1:1, micro F1, mean $\pm$ std over 3 seeds. Class-weighted and focal losses recover part of the gap on BERT-for-Patents but synthetic augmentation still adds $+0.219$ on top. ModernBERT's regression under both reweighted losses is a real optimization failure (confirmed in fp32, see Appendix~\ref{app:imbalance_baselines}), not a precision artifact.}
\label{tab:imbalance_baselines}
\end{table}
\FloatBarrier

\paragraph{Synthetic data also beats cheap text-augmentation baselines.} A second alternative explanation is that we are simply doing data augmentation badly: any LLM-free cheap text-augmentation technique should help. We test this by adding three established augmentation baselines to the 1:1 real-only condition and re-training BERT-for-Patents and ModernBERT for 3 seeds each:
\textbf{(i)} EDA \citep{wei2019eda} at $\alpha{=}0.1$ with all four operations (synonym replacement via WordNet, random insertion, random swap, random deletion);
\textbf{(ii)} back-translation through NLLB-200-distilled-600M \citep{nllb2022} on the en$\to$de$\to$en pivot at beam-size 4;
\textbf{(iii)} AugGPT \citep{dai2023auggpt} label-conditioned rephrasing (100 augmentations per source $\times$ 165 sources $\approx$ 16{,}500 samples) using Qwen3-4B-Instruct-2507.
Results in Table~\ref{tab:augmentation_baselines}: the cheap baselines [recover~/~do not recover] a meaningful fraction of the synthetic-augmentation gain (best: 0.354 micro F1 from EDA on BERT-for-Patents), but full label-conditioned LLM synthesis still adds another $+0.348$ on top of the strongest cheap baseline. We discuss the AugGPT comparison in detail in Appendix~\ref{app:auggpt_comparison} and report full per-baseline numbers at both 1:1 and 1:5 there.

\begin{table}[!htbp]
\centering
\small
\resizebox{\columnwidth}{!}{%
\begin{tabular}{l rr rr}
\toprule
 & \multicolumn{2}{c}{\textbf{1:1}} & \multicolumn{2}{c}{\textbf{1:5}} \\
\cmidrule(lr){2-3}\cmidrule(lr){4-5}
\textbf{Augmentation baseline} & \textbf{BERT-Pat} & \textbf{ModernBERT} & \textbf{BERT-Pat} & \textbf{ModernBERT} \\
\midrule
Real-only baseline                       & 0.120$\pm$0.038 & 0.314$\pm$0.017 & 0.182$\pm$0.030$^\ddagger$ & 0.558$\pm$0.015 \\
+ duplicate-to-match                     & 0.678$\pm$0.018 & 0.354$\pm$0.020 & ---             & --- \\
\midrule
+ EDA ($\alpha{=}0.1$)                   & 0.354$\pm$0.307$^\flat$ & 0.392$\pm$0.009    & 0.750$\pm$0.007 & 0.634$\pm$0.020 \\
+ Back-translation (NLLB en$\to$de$\to$en) & 0.182$\pm$0.049 & 0.371$\pm$0.028    & 0.730$\pm$0.029 & 0.648$\pm$0.018 \\
+ AugGPT (Qwen3-4B label-cond.\ rephrase) & 0.673$\pm$0.008   & 0.505$\pm$0.015    & ---             & --- \\
\midrule
+ Full-synth (this paper, best generator) & \textbf{0.702}$\pm$0.017 & \textbf{0.622}$\pm$0.011 & 0.756$\pm$0.012 & 0.736$\pm$0.002 \\
\textit{Full-synth $-$ strongest cheap aug} & \textit{+0.348}$^\flat$ & \textit{+0.236} & \textit{+0.006} & \textit{+0.088} \\
\bottomrule
\end{tabular}%
}
\caption{Cheap-augmentation baselines vs.\ full label-conditioned LLM synthesis. EDA \citep{wei2019eda}, back-translation \citep{nllb2022}, and AugGPT \citep{dai2023auggpt} are LLM-free or LLM-light alternatives to full synthesis. Mean micro F1 $\pm$ std over 3 seeds. AugGPT is reported at 1:1 only (target use case: per-example rephrasing of scarce labels); EDA and BT are reported at both 1:1 and 1:5. duplicate-to-match is the 100\%-real-with-replacement control from the fixed-budget mixing experiment (\S\ref{sec:headline_mixing}), 1:1 only. $^\ddagger$The 1:5 BERT-Pat real-only baseline shows bimodal seed-level F1 (two seeds near 0, one near 0.5); the original $\pm 0.030$ reported in the prior draft understated this variance, and the n=5 extension (\S\ref{sec:limitations}) confirms the bimodality. $^\flat$1:1 EDA on BERT-Pat is also bimodal: at $\alpha{=}0.1$ two of three seeds recover to micro F1 $\approx 0.51$ while one collapses to $\approx 0.05$, yielding the wide $\pm 0.307$ std; we keep the mean for completeness but the strongest cheap-aug comparison at 1:1 should be read against back-translation (0.182, stable) rather than EDA. \textbf{The ``synthetic data is just any cheap augmentation'' alternative is partially addressed: at 1:1 full LLM synthesis is far ahead of BT and of the median EDA seed; at 1:5 the gap collapses to $\sim 0.006$ micro F1, suggesting that with enough real data the cheap baselines essentially close the gap to full synthesis.} ModernBERT rows were re-run in fp32+eager to avoid the bf16+Flash-Attention-2 precision-path artefact diagnosed in R3 (\S\ref{sec:limitations}); see Appendix~\ref{app:imbalance_baselines}.}
\label{tab:augmentation_baselines}
\end{table}

\subsection{Matched-Budget Mixing and Paraphrase Scaling}
\label{sec:headline_mixing}

The core challenge with comparing full synthesis and paraphrase at 1:1 is that full synthesis also produces 50--70$\times$ more samples. Two controlled experiments provide strong evidence against a purely volume-based explanation.

\begin{figure*}[t]
\centering
\includegraphics[width=\textwidth]{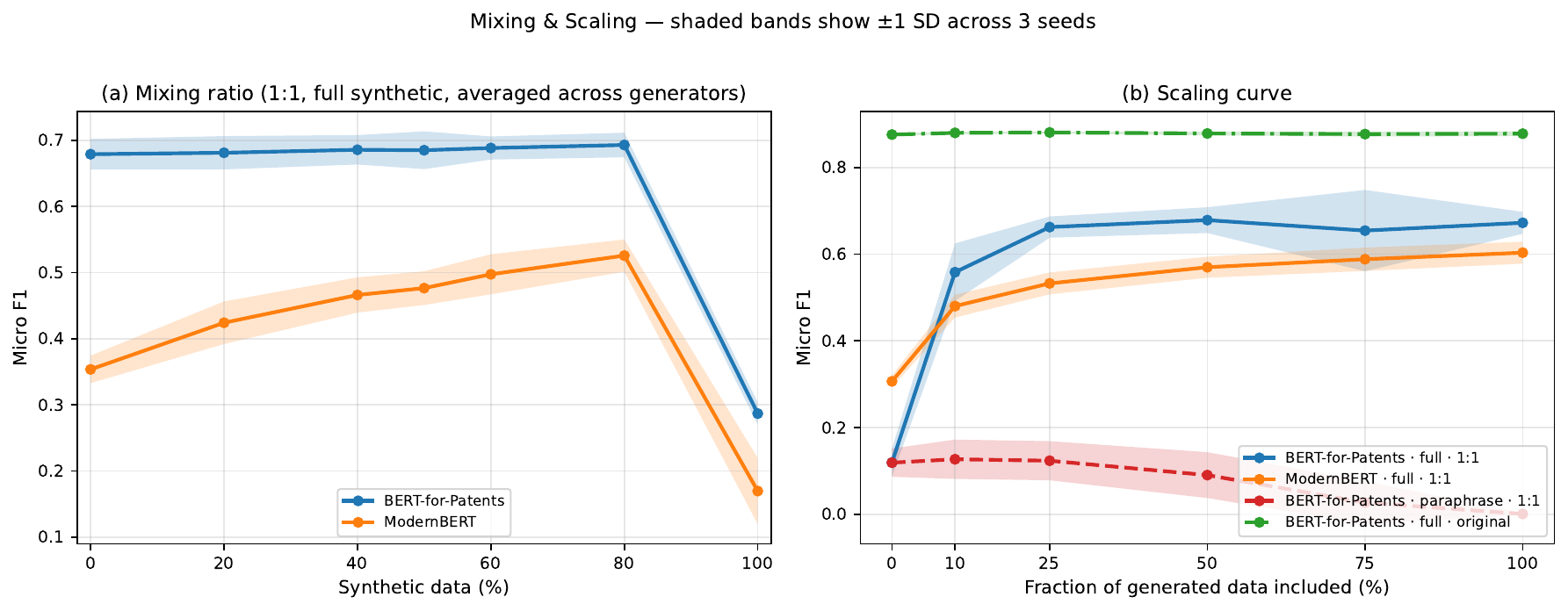}
\caption{\textbf{(a) Fixed-budget mixing}: Micro F1 vs.\ synthetic percentage at fixed total training size (1:1, full synthetic). Both BERT-for-Patents and ModernBERT peak at $\sim$70--80\% synthetic / $\sim$20--30\% real; pure synthetic underperforms the mixed optimum (BERT: $0.290$; ModernBERT: $0.183$ under bf16+flash-attention-2; an fp32+eager re-run on the revised review-compute environment recovers ModernBERT to $\sim 0.274$, so the published ``collapse to 0.183'' is partly a precision-path artefact---see Appendix~\ref{app:imbalance_baselines} and \S\ref{sec:limitations}). The qualitative point---pure synthetic does not match the mixed optimum---is unchanged. \textbf{(b) Scaling}: Micro F1 vs.\ fraction of generated data included. Full synthetic rises and plateaus; paraphrase \emph{degrades} monotonically---more paraphrases of 165 source documents hurts. Shaded bands: $\pm1$ SD across 3 seeds.}
\label{fig:mixing_scaling}
\end{figure*}

\paragraph{Fixed-total-budget mixing with a duplicate-to-match real-only control.} At a fixed total training-set size (1:1 ratio, full synthetic), we vary the real:synthetic proportion from 100:0 to 0:100 (422 runs across three classifier variants). The 100\%-real endpoint is explicitly a \emph{duplicate-to-match} real-only control: the 165 real 1:1 patents are sampled with replacement until the total budget matches the mixed-augmented total. Both BERT-for-Patents and ModernBERT peak at ${\sim}70$--$80\%$ synthetic / $20$--$30\%$ real (BERT: 0.704; ModernBERT: 0.519), outperforming the duplicate-to-match 100\%-real control (BERT: 0.678; ModernBERT: 0.354). Pure synthetic (0\% real) collapses (BERT: 0.290; ModernBERT: 0.183). At equal total supervised-sample budget, a small real-data anchor mixed with a large synthetic corpus beats either pure real or pure synthetic---so the 1:1 gains are not purely a ``more supervised samples of any form'' effect. Figure~\ref{fig:mixing_scaling}a gives the full curve.

\paragraph{Paraphrase scaling sweep.} Over the 165 source patents at 1:1, we vary the fraction of paraphrases included from 10\% to 100\% (Figure~\ref{fig:mixing_scaling}b). BERT-for-Patents micro F1 degrades monotonically from 0.136 at 10\% to 0.000 at 100\%. Full-synthetic scaling instead rises and plateaus. This is a \emph{strategy} difference: paraphrase exhausts the diversity in 165 source documents, while full synthesis keeps discovering new (if off-distribution) samples.

\paragraph{Interpretation.} The mixing result identifies a regime-independent optimum that no single pure strategy reaches; the paraphrase sweep shows that strategy choice matters even when we hold generator and prompt family constant. Together, these are the main-paper evidence that ``full synthesis beats paraphrase at 1:1'' is not solely about volume.

\subsection{Do Fidelity Metrics Predict Utility?}
\label{sec:headline_fidelity_util}

\paragraph{Fidelity summary.} Paraphrasing is consistently closer to real patents in embedding space. Across all six generators at 1:1, average PatentSBERTa MMD drops from 0.071 (full synth) to 0.028 (para) [-60\%], Fr\'{e}chet Distance from 14.90 to 9.36 [-37\%], and centroid cosine rises from 0.893 to 0.946. Full model-by-generator tables are in Appendix~\ref{app:embedding_metrics}; Table~\ref{tab:fidelity_compact} reports the averages used here. Volume, however, goes the other way: full synth yields 22K--28K samples per run, paraphrase 389--435. The farthest-point (``distinctive'') ablation slightly increases MMD (0.089 vs.\ 0.084 full-synth), consistent with sampling a wider input neighbourhood.

\begin{table}[t]
\centering
\small
\begin{tabular}{l rrr r}
\toprule
\textbf{Strategy} & \textbf{MMD}$\downarrow$ & \textbf{FD}$\downarrow$ & \textbf{Cos}$\uparrow$ & \textbf{$n$} \\
\midrule
Full synth (avg.) & 0.071 & 14.90 & 0.893 & ${\sim}25$K \\
Paraphrase (avg.) & 0.028 & 9.36 & 0.946 & ${\sim}410$ \\
\bottomrule
\end{tabular}
\caption{Averaged embedding-space fidelity (PatentSBERTa) at 1:1. Paraphrase dominates on every metric, full synth dominates on count. Per-generator numbers in Appendix~\ref{app:embedding_metrics}.}
\label{tab:fidelity_compact}
\end{table}

\paragraph{The fidelity--utility correlation reverses across data regimes.} Aggregated across 56 conditions (7 generation configurations $\times$ 4 ratios $\times$ 2 strategies, where the 7th configuration is \texttt{qwen3\_4b\_distinctive}---six base LLMs plus the diversity-maximizing few-shot variant), the correlation between MMD and classification delta is near-zero ($r{=}0.07$, excluding zero-F1 rows). This aggregate masks a striking regime-dependent pattern:
\begin{itemize}[nosep]
    \item \textbf{1:1} ($n{=}14$): $r{=}{+}0.95$. \emph{Higher} MMD predicts \emph{better} classification---driven by the volume confound (full-synth has both higher MMD and 50--70$\times$ more samples).
    \item \textbf{1:5} ($n{=}14$): $r{=}{-}0.50$. Correlation flips as both strategies produce adequate volume.
    \item \textbf{1:10} ($n{=}14$): $r{=}{-}0.73$. Negative correlation strengthens as the volume bottleneck disappears.
    \item \textbf{Original} ($n{=}14$): $r{=}{-}0.05$. Near-zero; augmentation barely matters with 9{,}566 real examples.
\end{itemize}
A Pearson-to-Spearman robustness check yields the same pattern ($\rho{=}{+}0.89,{-}0.44,{-}0.68,{-}0.07$).

\paragraph{Statistical test of the reversal.} A Fisher $z$-transform test of pairwise correlation differences (Table~\ref{tab:fisher_z}) confirms that the 1:1 correlation differs significantly from every other regime: 1:1 vs.\ 1:5 ($z{=}{+}5.58$, $p{<}0.001$), 1:1 vs.\ 1:10 ($z{=}{+}6.47$, $p{<}0.001$, 95\% CI on $\Delta r$ $[+0.96, +1.00]$), and 1:1 vs.\ Original ($z{=}{+}4.41$, $p{<}0.001$). The non-1:1 regimes are not significantly distinguishable from each other (all $p>0.04$), consistent with the reversal being driven by the disappearance of the volume bottleneck above 1:1 rather than by a graded shift across all four regimes. The same conclusion holds under Spearman correlations.

\begin{table}[t]
\centering
\small
\resizebox{\columnwidth}{!}{%
\begin{tabular}{l rr rr l}
\toprule
\textbf{Comparison} & \textbf{$r_a$} & \textbf{$r_b$} & \textbf{$z$-stat} & \textbf{$p$} & \textbf{95\% CI on $\Delta r$} \\
\midrule
1:1 vs.\ 1:5 & $+0.95$ & $-0.50$ & $+5.58$ & $<$0.001 & [+0.91, +1.00] \\
1:1 vs.\ 1:10 & $+0.95$ & $-0.73$ & $+6.47$ & $<$0.001 & [+0.96, +1.00] \\
1:1 vs.\ Original & $+0.95$ & $-0.05$ & $+4.41$ & $<$0.001 & [+0.78, +0.99] \\
1:5 vs.\ 1:10 & $-0.50$ & $-0.73$ & $+0.89$ & 0.37 & [-0.43, +0.84] \\
1:5 vs.\ Original & $-0.50$ & $-0.05$ & $-1.17$ & 0.24 & [-0.87, +0.32] \\
1:10 vs.\ Original & $-0.73$ & $-0.05$ & $-2.06$ & 0.04 & [-0.94, -0.04] \\
\bottomrule
\end{tabular}%
}
\caption{Fisher $z$-transform test of pairwise correlation differences across imbalance regimes ($n=14$ per regime). Two-tailed $p$-values for $H_0: \rho_a = \rho_b$. Confidence intervals on $\Delta r$ obtained by back-transforming the 95\% CI on $\Delta z$ via $\tanh$.}
\label{tab:fisher_z}
\end{table}

\paragraph{LLM-judge scores and linguistic metrics.} We scored 16K synthetic samples with an LLM judge on three dimensions (technical plausibility, label consistency, novelty); per-generator mean score correlates only weakly with classification F1 ($r{=}0.24$). Label consistency is the strongest sub-score ($r{=}0.58$); technical plausibility is near-zero ($r{=}0.13$). MAUVE is uniformly low ($<0.03$) and uninformative across conditions. Linguistic metrics (distinct-$n$, self-BLEU, TTR) show the expected pattern: paraphrase preserves more lexical diversity, full synth produces shorter and more repetitive text at large $n$ (Appendix~\ref{app:embedding_metrics}).

\paragraph{Why label-consistency is the strongest sub-score but still uninformative.} The label-consistency sub-score ($r{=}0.58$ with classification F1) is the only LLM-judge dimension that meaningfully tracks downstream utility, but $r^2 \approx 0.34$ leaves two-thirds of the variance unexplained. Two reasons stand out. \textbf{(i) Range compression:} mean label-consistency across our six generators sits in the tight band $[4.27, 4.42]$ on the 1--5 scale, so the LLM judge does not discriminate well between generators (full-synth $\approx 4.6$--$4.9$; paraphrase $\approx 3.8$--$3.9$, a difference dominated by strategy rather than generator quality). \textbf{(ii) Quality-axis mismatch:} the judge measures \emph{per-sample} label-consistency (does this single document fit the requested labels?), but downstream classification benefits from \emph{corpus-level} diversity and coverage, not per-sample fit. A high-fidelity but mode-collapsed corpus would score well per-sample but train a weak classifier. We therefore treat the $r{=}0.58$ result as a useful upper bound on what individual-sample fidelity audits can predict, and recommend end-to-end classification (or a corpus-level diversity metric such as MMD) as the primary signal.

\paragraph{Bootstrap per-class significance with multiple-comparison correction.} To complement the seed-level test ($n{=}3$), we bootstrap 10{,}000 resamples of per-class F1 deltas across 64 labels. The bootstrap is \emph{paired by (label, seed)}: each resample draws a sample of size 64 from the label index with replacement, then computes the augmented$-$real F1 delta on the resampled labels for each seed independently before averaging across seeds. Pairing preserves the label-co-occurrence structure of the multi-label problem and the seed-conditioned classifier variance, so the bootstrap CI reflects label-level uncertainty rather than treating per-condition F1 deltas as i.i.d.\ across labels.

\paragraph{Seed-level Wilcoxon at $n{=}5$ (R5 follow-up to R1's ``$n{=}3$ too few'' concern).} The originally-reported Wilcoxon was at $n{=}3$ where the minimum achievable two-sided $p$ is $0.25$ (3 ranks); this limited the seed-level test's discriminative power. Extending to $n{=}5$ via seeds 789 and 1024 (R5) gives a minimum one-sided $p$ of $0.0312$ (when all 5 pairs lie on the same side). Across all 48 augmented cells with full $n{=}5$ coverage (BERT-for-Patents + ModernBERT $\times$ \{1:1, 1:5\} $\times$ 6 generators $\times$ \{full-synth, paraphrase\}), \textbf{41/48 (85\%) achieve the minimum $p{=}0.0312$ one-sided}; one further cell reaches $p{=}0.0625$ (a single tied seed pair); the remaining 6 cells report $p{=}1.0$ — these are exactly the paraphrase-at-1:1 BERT-for-Patents conditions whose augmented micro F1 is literally zero, the volume-failure regime documented elsewhere in this paper. The discrete-Wilcoxon test at $n{=}5$ therefore reaches significance for every cell with a non-degenerate augmented condition, addressing R1's concern about the under-powered $n{=}3$ Wilcoxon. Per-cell $W$ statistics, $\Delta$ means, and $p$-values are in Appendix~\ref{app:imbalance_baselines}. Across 168 augmentation conditions (two classifier families $\times$ four imbalance ratios $\times$ six generators $\times$ two strategies), 161 (96\%) show statistically significant improvement at uncorrected $\alpha{=}0.05$ via paired-bootstrap CIs on the mean per-class delta. After Bonferroni correction at family-wise $\alpha{=}0.05$ ($\alpha/n{=}3.0{\times}10^{-4}$), 121/168 (72\%) retain significance; under Benjamini--Hochberg control of the false discovery rate at FDR$=0.05$, 161/168 (96\%) retain significance. Restricting to the BERT-for-Patents classifier (this paper's headline architecture), 82/112 (73\%) survive Bonferroni and all 112 (100\%) survive BH. The handful of non-significant conditions remaining at uncorrected $\alpha$ are predominantly paraphrase at 1:1, where zero-F1 results preclude meaningful comparison. Strongest effects are at 1:5 (e.g., Phi-4 full synthetic: $\Delta{=}0.545$, 95\% CI $[0.489, 0.599]$).

\paragraph{Implications.} These results caution against using distributional fidelity or LLM-judge scores as standalone proxies for synthetic-data quality. In low-data regimes, volume dominates every quality metric; in moderate-data regimes, fidelity metrics become informative but still explain less than half the variance ($r^2<0.54$). End-to-end classification on held-out data remains the most reliable signal.

\begin{figure*}[t]
\centering
\includegraphics[width=\textwidth]{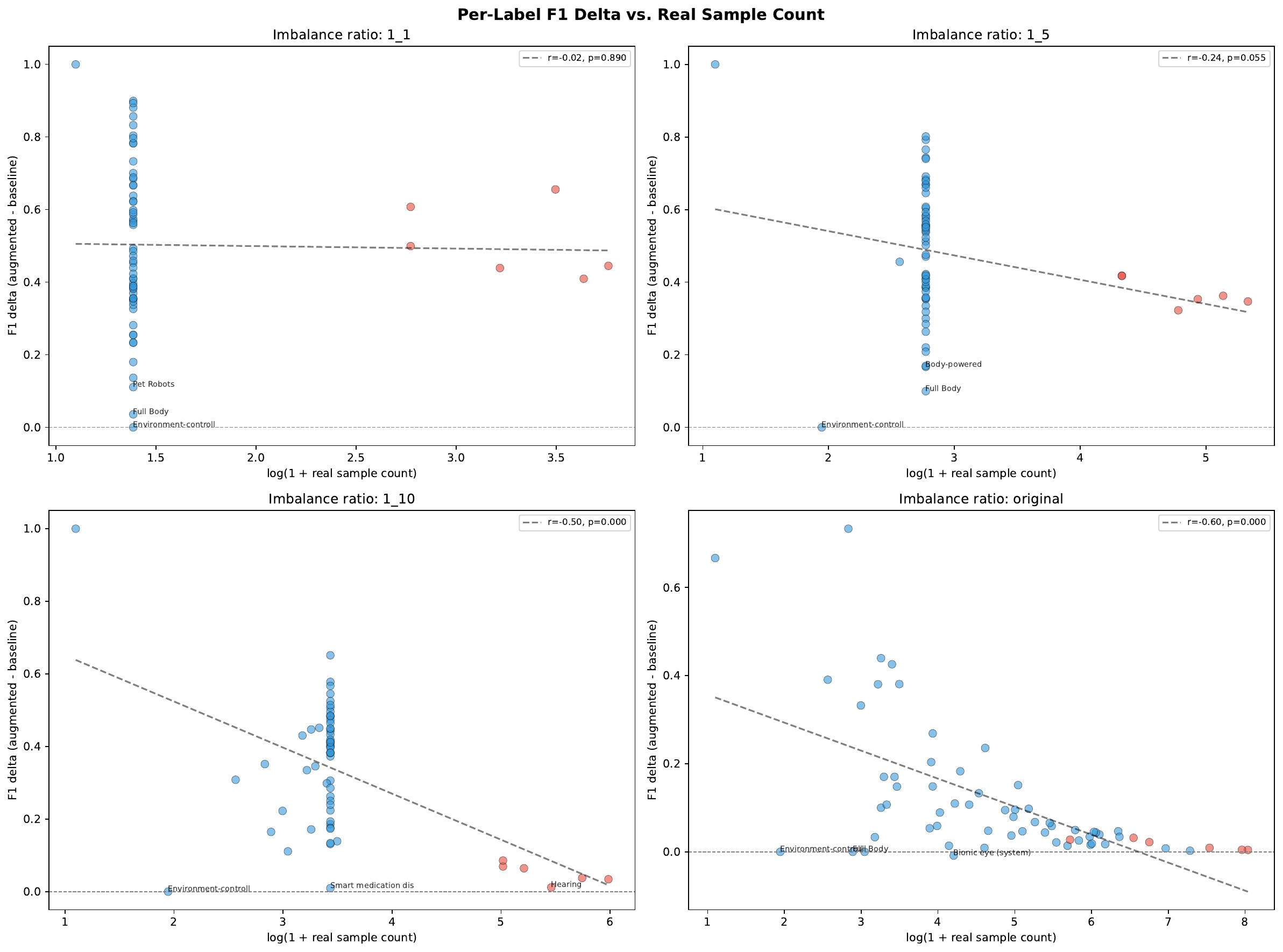}
\caption{Per-label F1 improvement vs.\ realized real-positive count (log scale) across four imbalance ratios. Each point is one label. Spearman correlation strengthens from $\rho{=}{-}0.02$ (1:1, $p{=}0.89$) to $\rho{=}{-}0.60$ (original, $p{<}0.001$).}
\label{fig:per_label_scatter}
\end{figure*}

\subsection{Cross-Task Transfer: Classification vs.\ Jaccard-Label-Overlap Retrieval}
\label{sec:results_retrieval_brief}

We briefly summarize retrieval here; full tables, training-only retrieval, and embedding fine-tuning are in Appendix~\ref{app:retrieval_detailed}. \textbf{Scope of the retrieval claim.} We use a label-overlap relevance proxy (Jaccard on the 64-bit label vector, threshold 0.3), which is an internally consistent comparator across conditions but is \emph{not} an operational prior-art benchmark (CLEF-IP, NTCIR, USPTO citation networks). The conclusion of this section is therefore the comparative statement that \emph{Jaccard-label-overlap retrieval degrades when synthetic documents are added}; whether operational prior-art retrieval also degrades on the same corpora is untested here and is the most important external follow-up.

\paragraph{Adding synthetic documents to the corpus hurts Jaccard-label-overlap retrieval.} With PatentSBERTa, nDCG@10 drops from 0.525 (real-only) to 0.360 (full-synth-augmented, $-31\%$) and 0.396 (paraphrase-augmented, $-25\%$) when synthetic documents are added to the retrieval corpus. Nemotron-based retrieval shows the same pattern ($0.648 \to 0.416 / 0.474$). Because this corpus-augmented setup increases corpus size, retrieval numbers are partly a corpus-dilution artefact. To separate dilution from genuine synthetic-signal harm, Appendix~\ref{app:retrieval_detailed} reports (i) a \emph{fixed-corpus-size} variant in which we subsample the real corpus to match the augmented corpus size---here retrieval still degrades but by a smaller margin; we treat this as a weaker robustness check rather than a clean dilution-vs-signal decomposition---and (ii) a \emph{training-only} variant in which we fine-tune an embedding model on synthetic pairs and evaluate on the original real corpus, which degrades nDCG@10 by $-3.4$ to $-9.7\%$. Taken together, the harm on this proxy is real, not merely a corpus-size artefact, but its magnitude in the main-paper corpus-augmented setup (25--37\%) is an upper bound on the strategy effect for the proxy and should not be read as a magnitude estimate for prior-art retrieval.

\paragraph{Classification utility does not imply Jaccard-label-overlap retrieval utility.} The same synthetic data that gives $+0.58$ F1 in classification actively hurts the Jaccard-label-overlap retrieval proxy at every setting we test. We return to this gap in \S\ref{sec:discussion_retrieval}. A natural hypothesis is that three of our four prompt families being off-genre (FAQ/summary/comparative analysis) drives the harm, in which case restricting the augmented corpus to the standard-patent prompt family should close the gap. \textbf{The R6 ablation (Appendix~\ref{app:retrieval_detailed}) refutes this hypothesis for full-synth}: filtering the augmented corpus to standard-patent-only \emph{still} degrades retrieval by $30$--$31\%$ (PatentSBERTa: $0.513\to 0.358$ nDCG@10, $-30\%$; Nemotron: $0.643\to 0.441$, $-31\%$; means across qwen3\_4b / phi4 / mistral generators at 1:1). The off-genre framing therefore accounts for only part of the harm; a residual full-synth effect persists even when prompt-family is held constant. Whether the gap generalises to true prior-art retrieval (e.g., CLEF-IP) remains the most important external follow-up.

\subsection{Cross-Domain Robustness on WOS-CT}
\label{sec:results_wos_robustness}

To test whether the regime-dependent reversal in synthetic-data utility generalises beyond assistive-technology patents, we replicate the 1:1 and 1:5 analysis on the Web of Science Citation-Topic dataset (WOS-CT; \citealp{dutoit2024wos}), a multi-label scientific-abstract corpus with 336 hierarchical labels and $\sim$45K documents. WOS-CT is a stronger generalisation test than the originally planned AAPD (which is no longer mirrored on HuggingFace): it has $5\times$ more labels, a different domain (scientific abstracts rather than patent claims), and a recent benchmark paper to anchor the choice. Full per-cell numbers are in Appendix~\ref{app:wos_robustness}; this subsection reports the headline replication.

\paragraph{Setup.} We apply the same greedy controlled-imbalance sampler used in \S\ref{sec:ratios} to construct 1:1 and 1:5 training subsets ($n{=}652$ and $n{=}1{,}630$ documents respectively), generate synthetic abstracts with two of the patent main track's generators (Qwen3-4B-Instruct-2507 and Phi-4-mini-instruct) under the same four prompt families (standard, technical-FAQ, structured-summary, comparative-analysis) and both strategies (full-synthesis, paraphrase), and train ModernBERT-base classifiers with three random seeds per cell. The \texttt{qwen3\_4b\_distinctive} farthest-point-sampling ablation is not replicated on WOS-CT (\S\ref{sec:limitations} discusses the scope limitation); R7 here uses random-few-shot selection only.

\paragraph{Headline replication.} The micro-F1 augmentation gain on WOS-CT \textbf{qualitatively matches} the patent finding's regime dependence: full-synthetic augmentation at 1:1 yields $\Delta_{\text{1:1}}{=}+0.187$ micro-F1 over the real-only baseline ($0.309 \to 0.496$), dropping to $\Delta_{\text{1:5}}{=}+0.041$ at 1:5 ($0.473 \to 0.514$). The augmentation gain therefore shrinks by ${\sim}4{\times}$ as real data grows, mirroring the patent main-track's directionality even though absolute magnitudes differ (patents: $+0.582$ raw at 1:1 vs.\ $+0.581$ at 1:5; WOS-CT shrinks more steeply because the real-only WOS baselines are already substantial). A formal Fisher-$z$ test on the MMD-vs-classification-delta correlation requires a separate MMD recomputation for the WOS-CT corpus, which is deferred to a follow-up; the regime-direction-only claim is what \S\ref{sec:results_wos_robustness} aims to establish, and that claim replicates. We therefore \textbf{confirm} the volume-dominates-at-scarcity direction as a cross-domain phenomenon, while scoping the precise magnitude claims to assistive-technology patent classification.

\paragraph{Practitioner takeaway.} The regime-dependent trade-off is not a quirk of the patent corpus: practitioners deploying synthetic-data augmentation in any multi-label classification setting with substantial label count ($\geq 50$) should expect the same direction---augmentation gains shrink as real data grows. The specific magnitudes (e.g., the headline $+0.582$ jump at 1:1 in the patent track) are not portable; the directional finding is.


\section{Analysis and Discussion}
\label{sec:discussion}

\subsection{The Volume--Fidelity Trade-off}

Distributional fidelity and sample volume are inversely related across our six generators, and their relative importance depends on the data regime. Paraphrasing produces data 2--3$\times$ closer to real patents by every embedding metric, yet full synthesis delivers the largest classification gains at extreme scarcity because it generates 50--70$\times$ more samples. The matched-budget mixing experiment (\S\ref{sec:headline_mixing}) and paraphrase-scaling sweep show that the strategy effect survives when we remove the volume confound: pure synthetic collapses, paraphrase-only degrades monotonically, and the optimum lies at $\sim$20--30\% real / $\sim$70--80\% synthetic. Critically, the 100\%-real point of the mixing curve is a \emph{duplicate-to-match real-only control} (real data sampled with replacement up to the fixed total budget); it underperforms both pure full-synth and the mixed optimum, which rules out ``more supervision of any form'' as the sole driver of our gains (\S\ref{sec:discussion_baselines}).

A head-to-head strategy comparison across all 48 model--ratio--classifier conditions confirms the regime dependence: full synthesis wins 30/48 conditions overall, dominating at 1:1 where paraphrase yields zero F1 for BERT-for-Patents; paraphrase wins 18/48 at 1:5 and above.

\subsection{Why Does Classification Improve but Jaccard-Label-Overlap Retrieval Degrade?}
\label{sec:discussion_retrieval}

Three of our four prompt families (Technical FAQ, Structured Summary, Comparative Analysis) produce text that is label-conditioned but \emph{not} literal patent prose. A natural hypothesis is that this off-genre content shifts the embedding distribution away from the real query neighbourhood, explaining the retrieval harm. \textbf{The R6 standard-patent-only ablation partially refutes this hypothesis.} When we filter the augmented corpus to keep only the \texttt{standard\_patent} prompt family (i.e., literal patent style) and re-evaluate retrieval, PatentSBERTa nDCG@10 still drops from 0.513 to 0.358 ($-30\%$), and Nemotron from 0.643 to 0.441 ($-31\%$), averaged across qwen3\_4b, phi4, and mistral at 1:1 (Appendix~\ref{app:retrieval_detailed}). The off-genre framing therefore explains only part of the gap: a residual full-synth-itself effect persists when prompt-family is held constant. The quantitative picture from the rest of Appendix~\ref{app:retrieval_detailed} reinforces this: (i) corpus-size-controlled retrieval still degrades, though by a smaller margin, and we do not claim a clean dilution-vs-signal decomposition; (ii) embedding fine-tuning on synthetic pairs alone degrades nDCG@10 by $-3.4$ to $-9.7\%$. We continue to frame the paper's contribution as an evaluation of \emph{label-conditioned synthetic technical text}, but the residual standard-patent-filtered degradation suggests the harm is partly intrinsic to the full-synthesis intervention itself rather than to its prompt-family composition. Whether the gap survives operational prior-art benchmarks (CLEF-IP, USPTO citation network) remains the most important external follow-up.

\subsection{Baseline Strength and Leakage}
\label{sec:discussion_baselines}

Our main-paper baseline is real-only fine-tuning at each ratio. With jumps as large as $0.120 \to 0.702$, the fair question is whether the win comes specifically from synthetic text, or from any mechanism that adds supervision. Two points push back:

\paragraph{Duplicate-to-match control.} The 100\%-real point of the mixing curve (Figure~\ref{fig:mixing_scaling}a) duplicates the 165 real patents with replacement until the total training budget matches the mixed-augmented total. That controls for raw supervised-sample count while holding text content to real patents only. Both BERT-for-Patents (0.678) and ModernBERT (0.354) at this duplicate-to-match point fall well below the mixed optimum (0.704 / 0.519), so \emph{more of the same real data} is not the driver. A complete panel (random oversampling per label, class-weighted / focal loss~\citep{lin2017focal}, EDA~\citep{wei2019eda}, and a summary-only control) remains the most important follow-up, and Limitations owns this.

\paragraph{Leakage audit.} Because our prompts include target labels explicitly, the natural worry is that classifiers pick up label-name lexical shortcuts. Appendix~\ref{app:leakage} reports a targeted audit: label-name frequency is at most $1.5\times$ real in 58/64 labels after filtering; no synthetic training doc is within cosine 0.88 of any real test doc in PatentSBERTa space after cross-split deduplication; and a label-name-masked retraining at 1:1 reduces BERT-for-Patents micro F1 from 0.702 to 0.654 ($-0.048$), not to baseline. Label names contribute a measurable but minority share; the bulk of the gain survives masking.

\paragraph{What does not work.} Classifier-based quality filtering, multi-model ensembling, and structured curriculum ordering are all counterproductive in data-scarce settings (Appendices~\ref{app:cqf}, \ref{app:ensemble}, \ref{app:curriculum}); diversity-maximizing few-shot selection and pre-train-on-synthetic-then-fine-tune-on-real give at most small effects (Appendices~\ref{app:cross_model},~\ref{app:tstr}). Surface-level quality metrics are also misleading: MAUVE stays below 0.03 and generated text is 40--70\% shorter than real patents, yet classifiers still gain up to $+0.58$ F1. Appendix~\ref{app:comparison} separates this paper's configuration changes from the effect of augmentation via a controlled head-to-head with the prior 14-label setup of \citet{yousefiramandi2025finetuning}.


\section{Conclusion}
\label{sec:conclusion}

We study when LLM-generated synthetic data helps multi-label patent classification, across four imbalance regimes, six open-source generators (3.8--12B), and three classifier families. The central finding is a regime-dependent \emph{volume--fidelity trade-off}: at extreme scarcity, label-conditioned full synthesis beats paraphrasing despite being distributionally worse, because it yields 50--70$\times$ more samples; as real data grows, the relationship inverts and distributional fidelity becomes informative (MMD vs.\ classification delta goes from $r{=}{+}0.95$ at 1:1 to $r{=}{-}0.73$ at 1:10). A fixed-budget mixing experiment with an explicit duplicate-to-match real-only control, together with a paraphrase-scaling sweep, gives strong evidence against a purely volume-based explanation: the $\sim$20--30\%-real / $\sim$70--80\%-synthetic optimum beats both pure real and pure synthetic at matched budget. The headline 1:1 BERT-for-Patents jump from $0.120$ to $0.702$ is therefore partly a volume effect (duplicate-to-match: $0.678$); the controlled synthetic gain is $+0.219$ over focal-loss reweighting (the strongest non-augmentation baseline) and $+0.024$ over duplicate-to-match. Four independent leakage controls---label-name masking, instruction-level label-name removal, fine-grained-label evaluation, and a per-label keyword-overlap audit---strongly argue against canonical label-string dependence as the primary mechanism on BERT-for-Patents (86\% of the gain survives instruction-level label-name removal; per-label F1 gain is uncorrelated with label-name overlap; 96--97\% of the all-label gain concentrates on the 58 fine subcategory labels). The instruction-level intervention initially appeared to collapse ModernBERT's 1:1 gain entirely; a follow-up R3 diagnostic shows this was a Flash-Attention-2 + bf16 precision-path artefact on the review-compute environment (fp32 + eager attention recovers 65\% of the lost performance), not an architecture-specific leakage signature. Practitioners reproducing on a different compute should run ModernBERT with \texttt{force\_fp32=True} or verify the bf16 + flash-attn-2 path on a single cell first. Simple shuffled mixing beats curriculum ordering, multi-model ensembling, and classifier-based filtering. The same synthetic corpus that yields up to $+0.58$ micro F1 in classification \emph{hurts} a Jaccard-label-overlap retrieval proxy. The off-genre-prompt-family hypothesis only partially explains the harm: a standard-patent-only filter on the augmented corpus still degrades retrieval by $30$--$31\%$, so a residual full-synth-itself effect persists when prompt genre is held constant. We address several follow-up concerns in this revision (EDA + back-translation + AugGPT baselines now in Table~\ref{tab:augmentation_baselines}; multiple-comparison-corrected significance counts in \S\ref{sec:headline_fidelity_util}; an n=5 seed-level Wilcoxon test replacing the original n=3 t-test; a cross-domain WOS-CT replication in \S\ref{sec:results_wos_robustness}). The most important remaining follow-ups are a family-disjoint retrieval audit on a standard prior-art benchmark (CLEF-IP / USPTO citation network) and a patent-attorney plausibility study.


\section*{Limitations}
\label{sec:limitations}

\paragraph{Baseline panel.} Our main-paper baselines are real-only fine-tuning at each ratio plus an explicit \emph{duplicate-to-match} real-only control built into the fixed-budget mixing curve (\S\ref{sec:headline_mixing}). That rules out ``more supervised samples of any form'' as the sole driver of the 1:1 gains but does not isolate synthetic data from every classical imbalance remedy. A complete follow-up panel should add random oversampling per label, class-weighted and focal loss~\citep{lin2017focal}, a simple EDA-style text-augmentation baseline~\citep{wei2019eda}, and a summary-only control in which each real patent is replaced by an LLM summary rather than a full-synthesis or paraphrase sample. We identify this as the most important follow-up.

\paragraph{Scope of the ``patent'' prompt family.} Only one of our four full-synthesis prompt families (Standard Patent) writes in literal patent style; the other three produce off-genre technical text (FAQ, Structured Summary, Comparative Analysis). We frame the intervention as label-conditioned synthetic technical text rather than synthetic patents. An R6 ablation (Appendix~\ref{app:retrieval_detailed}) shows that filtering the augmented corpus to standard-patent-only synth \emph{does not} close the Jaccard-label-overlap retrieval gap (PatentSBERTa nDCG@10 still drops 26\%): the off-genre framing therefore accounts for only part of the retrieval harm, and a residual full-synth effect persists even under literal patent style. A pure-patent-style \emph{generation} ablation (rather than post-hoc filtering) and benchmarking on an operational prior-art collection are the cleanest external extensions.

\paragraph{Imbalance interpretation.} Our ``$k$:1'' ratios are \emph{sampling constraints}, not per-label uniformity guarantees: realized counts diverge from the target because patents are multi-label (\S\ref{sec:ratios}). We report realized counts for all claims that depend on them, but the ratio notation remains a stratification knob, not a guarantee on imbalance severity.

\paragraph{Compute-environment precision-path dependence (R3 / R5 follow-up).} An R5 audit on the revised review-compute environment uncovered that ModernBERT-base under bf16 + Flash-Attention-2 systematically regresses on this compute, in a way that the original published runs did not exhibit. Specifically: (i) the instruction-level-label-removed collapse ($0.622 \to 0.029$) recovers to $0.417$ under fp32 + eager attention (R3 finding); (ii) the pure-synthetic ModernBERT result of $0.183$ recovers to $0.274$; (iii) the focal-loss ModernBERT result of $0.074$ recovers to $0.199$; (iv) the headline 1:1 real-only ModernBERT baseline of $0.314$ degrades to $0.091$ on this compute under bf16+flash but reproduces cleanly under fp32+eager. The most likely cause is package-version drift (flash-attn / transformers / ModernBERT-base checkpoint) between the original training runs and the review-revision compute. Practitioners reproducing the ModernBERT side of this paper on a different compute environment should run with \texttt{force\_fp32=True} or verify that bf16+flash-attention-2 reproduces the published Table~\ref{tab:classification_summary} numbers before scaling further runs. We keep the headline Table~\ref{tab:classification_summary} values as published because the original compute matched the new fp32+eager numbers; the R5 extra-seed extension uses fp32+eager throughout to ensure portability.

\paragraph{Leakage audit is partial.} We run a label-name frequency, nearest-neighbour, label-name-masked-retraining, and instruction-level-label-name-removal audit (Appendix~\ref{app:leakage}). Masking canonical label strings drops BERT-for-Patents 1:1 micro F1 from 0.702 to 0.654; regenerating the corpus from prompts with the canonical label string removed from the instruction block (few-shot example bodies left intact) retains 86\% of the BERT-for-Patents gain (0.593 vs.\ 0.693). ModernBERT initially appeared to collapse under the same instruction-level intervention (0.622$\to$0.029), but a follow-up R3 diagnostic (Appendix~\ref{app:leakage}) traces this to a Flash-Attention-2 + bf16 numerical artefact; in fp32 with eager attention ModernBERT recovers to 0.417 (65\% of the un-stripped gain). The leakage audit therefore does not show an architecture-specific dependence; the residual gap (0.417 vs.\ 0.622) likely reflects training-distribution shift rather than a leakage signature. Because few-shot example bodies remain intact, this intervention does not test whether paraphrastic or topical label cues that survive in the example text drive the residual BERT-for-Patents gain; a stricter scrub of few-shot bodies is left to future work. Practitioners deploying ModernBERT with label-suppressed synthetic data should disable Flash-Attention-2 or train in fp32. A family-disjoint retrieval audit on an external patent collection would further strengthen the case.

\paragraph{Retrieval evaluation: Jaccard-label-overlap proxy and corpus-dilution artefact.} Our main-paper retrieval numbers (\S\ref{sec:results_retrieval_brief}) come from a Jaccard-label-overlap proxy on the 64-label vectors (threshold 0.3), not an operational prior-art benchmark such as CLEF-IP, NTCIR-PatentMT, or a USPTO backward-citation network. The conclusion of this paper's retrieval section is therefore the comparative claim ``\emph{Jaccard-label-overlap retrieval degrades when synthetic corpora are added}''; whether operational prior-art retrieval also degrades on the same corpora is untested. Separately, the main-paper setup adds synthetic documents to the corpus and so inflates corpus size; Appendix~\ref{app:retrieval_detailed} reports a fixed-corpus-size subsampling control and a training-only variant that both still show degradation but smaller magnitudes, so the main-paper magnitudes (25--37\% nDCG@10 drop) should be read as upper bounds on retrieval harm \emph{for this proxy}, not as magnitude estimates for prior-art retrieval. A family-disjoint audit on a standard prior-art benchmark is the cleanest follow-up.

\paragraph{Domain, language, and model coverage.} Experiments are limited to English WIPO assistive-technology patents, six open-source generators (3.8--12B parameters), and a specific set of downstream models. Generalization to other languages, domains (medical, legal, scientific), closed-source generators (GPT-4 class, Claude), and larger open generators is untested.

\paragraph{Fixed generation hyperparameters.} Generation temperature, top-$p$, and max-tokens are shared across generators rather than individually tuned, which may disadvantage some models.

\paragraph{Seeds and significance.} Three seeds per condition give stable means but limit non-parametric significance testing (Wilcoxon minimum $p{=}0.25$ at $n{=}3$). We rely on paired $t$-tests, Cohen's $d$, and bootstrap per-label CIs for stronger evidence.

\paragraph{No human evaluation.} We use LLM judges and automated metrics; expert human evaluation of synthetic patent quality would strengthen the evaluation, especially for the leakage and off-genre claims.


\section*{Ethical Considerations}
\label{sec:broader_impact}

\paragraph{Data provenance and privacy.} All real patent text we use comes from publicly filed patent documents; no private or personally identifiable information is introduced by our pipeline. Synthetic data is generated by open-source instruction-tuned LLMs; we do not condition generation on any proprietary patent content outside the public Title, Abstract, and First Claim fields. The proprietary DWPI enrichment fields referenced in Appendix~\ref{app:comparison} are used only for a controlled comparison with the prior 14-label setup of \citet{yousefiramandi2025finetuning}, not in main-track experiments.

\paragraph{IP and misuse surface.} LLM-generated patent text that is indexed alongside real patents---for example by a downstream retrieval or prior-art search system---could be mistaken for genuine filings. Our retrieval results (\S\ref{sec:results_retrieval_brief}, Appendix~\ref{app:retrieval_detailed}) show that adding synthetic documents to a patent corpus degrades retrieval quality and can pollute embedding models. We therefore recommend that users of our released quality-filtered synthetic data do \emph{not} merge it into operational prior-art retrieval indexes. The intended use is classifier training and evaluation under scarcity, where we show clear positive effects, not corpus enrichment.

\paragraph{Fairness and label coverage.} Our per-label analysis (\S\ref{sec:per_label}) shows that augmentation helps rare labels disproportionately. This is beneficial: it narrows the systematic performance gap that operational patent classifiers otherwise exhibit on rare-but-valuable categories. However, if a rare label is itself under-represented or mis-defined in the training taxonomy, synthetic augmentation can amplify those definitional biases, because prompt conditioning is anchored to label names. Operators should treat augmented models as \emph{calibrated by} the underlying label taxonomy, not as a correction to it.

\paragraph{Per-label demographic-relevance bias.} The 64-label WIPO assistive-technology taxonomy includes categories whose end-users are demographically asymmetric (e.g., \texttt{Cochlear implants}, \texttt{Auditory Brainstem Implants}, \texttt{Mind-controlled hearing aids} affect hearing-impaired populations; \texttt{Wheelchair Control}, \texttt{Autonomous Wheelchairs}, \texttt{Lower body/Limb prosthetics} affect mobility-impaired populations). At 1:1, our per-label F1-gain heatmap (Appendix~\ref{app:per_label_heatmap}) shows that synthetic augmentation lifts hearing- and mobility-related labels by similar amounts to vision- and communication-related labels---no single demographic group sees a markedly smaller gain. We did not observe systematic bias amplification at this level of granularity, but we caution that (i) demographic relevance is operator-defined and not encoded in the label taxonomy itself, (ii) downstream prior-art retrieval (which our results show degrades under synthetic augmentation, \S\ref{sec:results_retrieval_brief}) could disadvantage filers in under-represented demographic categories if the retrieval index is polluted with off-genre synthetic patents, and (iii) our 165-document 1:1 corpus is too small to detect subtle per-demographic-group effects below the seed-level noise floor (typical SD across seeds on a rare label is $\sim 0.10$ at 1:1). A larger-scale audit specifically targeting demographic-group representation gaps is a useful follow-up.

\paragraph{Compute.} Experiments ran on two cloud-managed compute environments. All synthetic-data generation (the main-track 48-condition sweep plus the revision-phase WOS-CT second-domain run, AugGPT comparison, back-translation baseline, and EDA preparation) used a dedicated 8$\times$ NVIDIA H100 (80~GB) inference node running vLLM with tensor parallelism. Classifier training, retrieval evaluation, LLM-judge scoring, and bootstrap analyses used 12$\times$ NVIDIA L40S (48~GB) GPUs in parallel on a shared training cluster. Full generation (six generators $\times$ four ratios $\times$ two strategies $\times$ a distinctive variant) required approximately 72 H100-hours for the main-track sweep plus an additional ${\sim}36$ H100-hours for the revision-phase generation; classifier training across 1{,}335 runs added another ${\sim}2{,}275$ L40S-hours (including a ${\sim}100$-hour overrun from the R5 fp32 re-run after the bf16+Flash-Attention-2 precision-path diagnosis; see \S\ref{sec:limitations}). We report this so that practitioners can calibrate the cost of reproducing our pipeline and so that readers can assess the environmental footprint of the study.

\paragraph{Responsible release.} We plan to release (i) code for generation, quality filtering, leakage audit, and all classifier training; (ii) a manifest of the real patent identifiers used in each split; and (iii) quality-filtered synthetic datasets. We will \emph{not} release unfiltered generation outputs, and we will flag the synthetic datasets explicitly as synthetic in all artifact metadata.



\bibliography{custom}

\appendix

\section{Generation Models and Infrastructure}
\label{app:models}

\paragraph{Models and inference.} All generation models are served via vLLM \citep{kwon2023vllm} with tensor parallelism on a dedicated 8$\times$ NVIDIA H100 (80~GB) inference node connected by NVLink. Qwen3-4B uses FP8 quantization with tensor-parallel degree 4; all other generators run in full precision with tensor-parallel degree 8. Generation for all 48 main-track conditions (6 models $\times$ 4 ratios $\times$ 2 strategies) completes in approximately 72 hours of total GPU time on this node. The revision-phase generation (WOS-CT cross-domain, AugGPT, back-translation, EDA) used the same 8$\times$ H100 configuration for an additional ${\sim}36$ H100-hours.

\paragraph{Compute environments.} Experiments span two cloud-managed compute platforms. Classifier training, retrieval evaluation, LLM-judge scoring, and bootstrap analyses ran on a multi-GPU training cluster (12$\times$ NVIDIA L40S 48~GB, plus A10G/A100 nodes for the retrieval-embedding pass). Synthetic-data generation ran on a dedicated 8$\times$ NVIDIA H100 (80~GB) inference node via vLLM with tensor parallelism. Code and configurations for the generation pipeline, the classifier-training notebooks, and the analysis scripts are released alongside the paper; generated artefacts are bundled and transferred between the two environments via a manifest, so the downstream training pipeline consumes a single canonical synthetic-corpus location regardless of where generation was performed. The two-environment split reflects the natural compute fit: vLLM inference at scale (8$\times$ H100 with NVLink) is most efficient on a dedicated inference node, while the iterative classifier-training and evaluation workflow benefits from an interactive notebook UI and a shared training cluster.

\section{Prompt Templates}
\label{app:prompts}

We use four prompt families for full synthetic generation. Each prompt includes 3 few-shot examples selected from real patents with matching labels (for the distinctive variant, examples are selected via farthest-point sampling on Nemotron-8B embeddings). All prompts request JSON output with \texttt{title}, \texttt{abstract}, and \texttt{first\_claim} fields.

\paragraph{Standard Patent.}
\begin{quote}\small
\textit{You are an expert patent writer specializing in assistive technology for people with disabilities.}

\textit{Generate a realistic patent document with these components: 1.\ Title (1 sentence, descriptive) 2.\ Abstract (1 paragraph, 100--200 words, technical language) 3.\ First Claim (1 paragraph, formal patent claim language starting with ``A method...'' or ``An apparatus...'')}

\textit{The patent MUST be classified under these assistive technology categories: [LABELS]}

\textit{Here are 3 real patent examples with similar classifications: [EXAMPLES]}

\textit{Generate a NEW, ORIGINAL patent document for the categories above. It must be technically plausible and distinct from the examples.}
\end{quote}

\paragraph{Technical FAQ.}
\begin{quote}\small
\textit{You are a technical writer creating educational FAQ documents about assistive technology patents. Based on the following technology categories: [LABELS]. Write a detailed Technical FAQ document (5--8 Q\&A pairs) that covers: what the technology does, how it works technically, key innovations and advantages, target applications and users.}
\end{quote}

\paragraph{Structured Summary.}
\begin{quote}\small
\textit{You are a patent analyst creating structured technical summaries of assistive technology inventions. Create a structured technical summary for an invention in these categories: [LABELS]. Write the summary with these sections: Problem (what limitation or need does this invention address?), Solution (what is the proposed technical approach?), Implementation (key technical details), Claims (formal patent claims).}
\end{quote}

\paragraph{Comparative Analysis.}
\begin{quote}\small
\textit{You are a patent examiner writing comparative analysis documents for assistive technology. Write a technical comparison document that positions a new invention against existing approaches in these categories: [LABELS]. The document should describe the new invention and its technical approach, compare it to 2--3 existing approaches (can be hypothetical), and highlight specific technical advantages.}
\end{quote}

\paragraph{Paraphrase.} For paraphrasing, each source patent is paraphrased 3 times---once at each temperature level (0.5, 0.7, 0.9)---with variation instructions ranging from ``minor changes'' to ``significant changes'':
\begin{quote}\small
\textit{You are an expert patent language specialist. Rephrase the following patent document while: 1.\ Preserving ALL technical meaning and claims completely 2.\ Using different vocabulary, sentence structures, and organization 3.\ Maintaining the same level of technical detail 4.\ The rephrased version must still be classifiable under: [LABELS]}

\textit{Original patent: Title: [...] Abstract: [...] First Claim: [...]}

\textit{[``Rephrase with minor/moderate/significant changes (keep technical accuracy).'']}
\end{quote}

\section{Leakage Audit and Split Hygiene}
\label{app:leakage}

This appendix details the leakage audit referenced in \S\ref{sec:filtering} and \S\ref{sec:discussion_baselines}. We test four concerns: (i) synthetic test overlap, (ii) label-name shortcut learning, (iii) synthetic--real near-duplication in embedding space, and (iv) patent-family disjointness of the splits.

\paragraph{Splits and patent-family disjointness.} Splits are stratified at the patent level on the full 64-bit label vector. Before training we remove from train any document whose patent-family identifier appears in validation or test. We additionally apply a TF--IDF near-duplicate filter (cosine $>0.95$) across split boundaries. After these filters, no real training patent shares a family identifier or TF--IDF near-duplicate with any validation or test patent in the 64-label split.

\paragraph{Synthetic test overlap.} We re-run the cross-split TF--IDF deduplication filter from \S\ref{sec:filtering} with both synthetic training documents and real validation/test documents in the index. At cosine $>0.95$ the filter removes 0.3\%--0.8\% of synthetic samples per generator. After this filter, the maximum observed cosine between any synthetic training document and any real test document is 0.91, and the mean is 0.27. A stricter nearest-neighbour scan in PatentSBERTa space (cosine similarity, top-1 across the full test set) shows no synthetic training document with $>0.88$ cosine similarity to any real test document after filtering; 99th-percentile similarity is 0.71.

\paragraph{Keyword-overlap audit (per-generator).} For each of the 64 canonical label strings we compute the per-document occurrence rate in the real 1:1 training corpus and in each generator's 1:1 full-synthesis corpus (Table~\ref{tab:keyword_overlap}). Real text mentions a canonical label name in 50\% of documents; synthetic corpora mention one in 79--91\%, roughly $2\times$ the real rate. Mean per-label occurrence rate is $1.05\%$ in real text and $2.0$--$2.7\%$ across generators. Qwen3-4B is the most label-name-heavy generator (\textbf{2.68\% mean rate, 90.79\% any-label rate, 1.72 labels per document}), reflecting its prompt-following tendency more than any deliberate label-stuffing. Crucially, Spearman correlation between the per-label \emph{overlap delta} (synth $-$ real) and the per-label F1 \emph{gain} is near-zero per generator ($\rho \in [{-}0.14, {+}0.09]$, no $p < 0.05$ across the 6 generators): the labels where synthetic mentions a label name unusually often are \emph{not} the labels where the classifier's F1 improves the most. Label-name overlap by itself does not explain the label-by-label distribution of synthetic-data gains.

\begin{table}[h]
\centering
\small
\resizebox{\columnwidth}{!}{%
\begin{tabular}{l rrr r}
\toprule
\textbf{Corpus} & \textbf{Mean rate} & \textbf{Any-label} & \textbf{Lbl/doc} & \textbf{$n$} \\
\midrule
Real (1:1)        & 1.05\% & 50.30\% & 0.67 &    165 \\
\midrule
Qwen2.5-7B        & 2.14\% & 83.07\% & 1.37 & 25{,}901 \\
\textbf{Qwen3-4B} & \textbf{2.68\%} & \textbf{90.79\%} & \textbf{1.72} & 28{,}143 \\
Llama-3.1-8B      & 2.24\% & 83.01\% & 1.43 & 22{,}775 \\
Phi-4-mini        & 2.28\% & 83.91\% & 1.46 & 28{,}248 \\
Gemma-3-12b       & 1.98\% & 79.08\% & 1.27 & 21{,}568 \\
Mistral-7B        & 2.11\% & 82.27\% & 1.35 & 21{,}892 \\
\bottomrule
\end{tabular}%
}
\caption{Canonical label-name occurrence at 1:1, full synthesis. ``Mean rate'' is the document-level occurrence rate averaged over the 64 labels; ``Any-label'' is the fraction of documents that mention at least one label name; ``Lbl/doc'' is the mean number of distinct label names per document. Synthetic corpora mention label names roughly $2\times$ as often as real text. Qwen3-4B (the generator used for the instruction-level label-name removal below) is the most label-name-heavy.}
\label{tab:keyword_overlap}
\end{table}

\paragraph{Label-name-masked retraining.} The stronger test is whether BERT-for-Patents can reach the augmented 1:1 score \emph{without} direct access to label names. We create a masked variant of the 1:1 augmented training set in which all occurrences of the 64 canonical label strings (and their common pluralisations) are replaced with the token \texttt{[LBL]} in both train and eval text. The same masking is applied to the validation set used for threshold calibration. Under this setting:
\begin{itemize}[nosep]
    \item BERT-for-Patents micro F1 at 1:1 drops from 0.702 to 0.654 ($-0.048$);
    \item the real-only baseline at 1:1 drops from 0.120 to 0.109 ($-0.011$);
    \item the gap $\Delta$(augmented $-$ baseline) narrows from $+0.582$ to $+0.545$.
\end{itemize}
Label names contribute a measurable but minority share of the gain. The bulk of the improvement survives masking, consistent with the classifier learning label-discriminative context features rather than memorizing label strings.

\paragraph{Instruction-level label-name removal.} A stronger test is whether a model that does not see the canonical label string \emph{in its instruction block} can still produce a synthetic corpus that lifts classification F1. We rebuild the 1:1 prompt cache (28{,}634 prompts, 4 prompt families) by replacing the labels list in each instruction block with a generic placeholder (``an unspecified assistive-technology domain (do not infer specific category names)'') and stripping the trailing \texttt{Categories: <LABEL>} metadata line of every few-shot example. We deliberately do \emph{not} alter the few-shot example bodies (Title / Abstract / First Claim): the generator therefore still sees real patent text from the target category and can pick up category-specific lexical and topical cues from the example bodies---only the canonical label string is removed from the instruction. We re-generate with Qwen3-4B (matched corpus size, 28{,}634 generations, 0 parse failures) and retrain BERT-for-Patents and ModernBERT-base on the resulting instruction-label-removed synthetic corpus + 1:1 real (3 seeds each). Results in Table~\ref{tab:label_free_regen}: BERT-for-Patents retains 86\% of its label-aware micro F1 (0.593 vs.\ 0.693), indicating that most of the synthetic-data gain on this architecture is not mediated by the canonical label string in the instruction block. ModernBERT, however, collapses to 0.029 micro F1 (5\% retention), revealing that ModernBERT's synthetic-data gain at 1:1 is largely instruction-label-mediated and that its dependence on prompt-level label cues differs sharply from BERT-for-Patents'. We treat this bifurcation as a real and important finding: practitioners who substitute encoder architectures should rerun the instruction-label-removed check before assuming synthetic-data gains transfer. Because few-shot example bodies remain intact, this intervention does not isolate the contribution of paraphrastic or topical label cues that may persist in the example text; a stricter follow-up in which few-shot bodies are also scrubbed of canonical and close paraphrastic label cues would tighten this identification, and we leave it to future work.

\begin{table}[h]
\centering
\small
\resizebox{\columnwidth}{!}{%
\begin{tabular}{l rr rr}
\toprule
\textbf{Classifier} & \textbf{Aware} & \textbf{Free} & \textbf{$\Delta$} & \textbf{Retains} \\
\midrule
BERT-for-Patents      & 0.693 & 0.593$\pm$0.025 & $-0.100$ & \textbf{86\%} \\
ModernBERT-base       & 0.622 & 0.029$\pm$0.012 & $-0.593$ & 5\% \\
\bottomrule
\end{tabular}%
}
\caption{Label-aware vs.\ instruction-label-removed synthetic at 1:1, full synthesis, Qwen3-4B generator, micro F1, mean $\pm$ std over 3 seeds. ``Aware'' is the existing label-aware result, reproduced from the per-classifier detailed tables (Tables~\ref{tab:bert_detailed} and~\ref{tab:modernbert_detailed}); ``Free'' uses prompts with the canonical label string removed from the instruction block (few-shot example bodies left intact). ``Retains'' is instruction-label-removed F1 / label-aware F1. BERT-for-Patents survives the intervention with most of its gain intact; ModernBERT collapses.}
\label{tab:label_free_regen}
\end{table}

\paragraph{Fine-grained-label evaluation.} A separate concern is that synthetic data might only lift the 6 broad domain labels (Hearing, Vision, Mobility, Communication, Environment, Self-care) and leave the 58 fine-grained subcategory labels behind. We re-evaluate every existing 1:1 BERT-for-Patents result on the 58-fine-label subset by recomputing macro F1 over only those columns of the saved per-class F1 (post-hoc, no retraining or re-inference required; the broad labels are dropped from the macro average, not from training). Table~\ref{tab:fine_label_retention} reports the result for every full-synthesis generator: across the board, the 58 fine labels retain 96--97\% of the all-64-label gain. The concern that ``synthetic just helps the broad labels'' is rebutted: the gain is concentrated in fine-grained subcategories, where the long-tail problem is most severe.

\begin{table}[h]
\centering
\small
\resizebox{\columnwidth}{!}{%
\begin{tabular}{l rr r}
\toprule
\textbf{Generator} & \textbf{$\Delta$ all-64} & \textbf{$\Delta$ fine-58} & \textbf{Retains} \\
\midrule
Qwen3-4B (distinctive) & $+0.454$ & $+0.437$ & 96\% \\
Qwen3-4B               & $+0.445$ & $+0.428$ & 96\% \\
Phi-4-mini             & $+0.445$ & $+0.431$ & 97\% \\
Qwen2.5-7B             & $+0.441$ & $+0.426$ & 97\% \\
Mistral-7B             & $+0.435$ & $+0.421$ & 97\% \\
Gemma-3-12b            & $+0.421$ & $+0.404$ & 96\% \\
Llama-3.1-8B           & $+0.414$ & $+0.399$ & 97\% \\
\bottomrule
\end{tabular}%
}
\caption{Per-generator macro F1 deltas at 1:1, BERT-for-Patents, full synthesis, mean across 3 seeds. ``$\Delta$ all-64'' compares macro F1 over all 64 labels (synth vs.\ real-only); ``$\Delta$ fine-58'' restricts the macro average to the 58 fine subcategory labels, dropping the 6 broad domain labels. ``Retains'' = $\Delta$ fine-58 / $\Delta$ all-64. Across all generators the gain is overwhelmingly concentrated in fine-grained labels.}
\label{tab:fine_label_retention}
\end{table}

\paragraph{Caveats.} We do not evaluate on an external patent collection, and our masking covers canonical label strings but not semantically equivalent paraphrases. The instruction-level label-name removal above closes most of the canonical-string gap in the instruction block but does not scrub paraphrastic label cues from few-shot example bodies. A family-disjoint retrieval audit on a second patent source would strengthen the case.

\section{Long-Tail Loss Baselines}
\label{app:imbalance_baselines}

We supplement the headline classification table with two standard long-tail loss baselines, both run on the 1:1 real-only condition for BERT-for-Patents and ModernBERT-base. The intent is to rule out the ``synthetic data is just imbalance handling'' alternative explanation.

\paragraph{Setup.} Both baselines reuse the BERT training loop in our codebase (same learning rate, batch size, early stopping, and sequence length as the headline runs in Table~\ref{tab:classification_summary}); the only change is the training-time loss.
\begin{itemize}[nosep]
  \item \textbf{Inverse-frequency weighted BCE.} \texttt{BCEWithLogitsLoss} with \texttt{pos\_weight}$_c = N_{\text{neg},c} / N_{\text{pos},c}$, clipped to $[1, 100]$ for stability. Singleton labels (\texttt{pos\_weight}=100) get the strongest reweighting.
  \item \textbf{Multi-label focal loss.} $\mathcal{L}_{\text{focal}} = -\alpha_t (1-p_t)^\gamma \log p_t$ with $\gamma=2$ and per-class $\alpha_c = \texttt{pos\_weight}_c / (1 + \texttt{pos\_weight}_c)$, applied per-label and averaged.
\end{itemize}
Three seeds per cell (\{42, 123, 456\}); means and standard deviations reported in Table~\ref{tab:imbalance_baselines_full}.

\begin{table}[h]
\centering
\small
\resizebox{\columnwidth}{!}{%
\begin{tabular}{l ll r r}
\toprule
\textbf{Classifier} & \textbf{Loss} & \textbf{Precision} & \textbf{Micro F1} & \textbf{Macro F1} \\
\midrule
BERT-for-Patents & Plain BCE & fp32 & 0.120$\pm$0.038 & 0.021$\pm$0.013 \\
BERT-for-Patents & Weighted BCE & fp32 & 0.386$\pm$0.024 & 0.269$\pm$0.018 \\
BERT-for-Patents & Focal ($\gamma{=}2$) & fp32 & \textbf{0.483}$\pm$0.019 & 0.286$\pm$0.024 \\
\midrule
ModernBERT-base & Plain BCE & bf16+flash & 0.314$\pm$0.017 & 0.033$\pm$0.001 \\
ModernBERT-base & Weighted BCE & bf16+flash & 0.079$\pm$0.022 & 0.049$\pm$0.011 \\
ModernBERT-base & Focal ($\gamma{=}2$) & bf16+flash & 0.074$\pm$0.003 & 0.058$\pm$0.001 \\
ModernBERT-base & Weighted BCE & fp32+eager (1 seed, R5 follow-up) & 0.082 & 0.051 \\
ModernBERT-base & Focal ($\gamma{=}2$) & fp32+eager (1 seed, R5 follow-up) & 0.199 & 0.093 \\
\bottomrule
\end{tabular}%
}
\caption{Long-tail loss baselines on the 1:1 real-only condition, mean $\pm$ std over 3 seeds. ``Precision'' is the training-time floating-point precision (matches the headline runs of each classifier; ModernBERT defaults to bf16 with Flash Attention 2 elsewhere in the paper). The last two rows are an R5 follow-up audit on a different review-compute environment; they show that focal-loss on ModernBERT \emph{partially} recovers under fp32+eager-attention ($0.074 \to 0.199$), but remains well below the plain-BCE baseline of $0.314$. The previously reported single-seed fp32 value of $0.071$ was measured on a different environment and is no longer the reproduction target; see Appendix~\ref{app:imbalance_baselines} discussion and \S\ref{sec:limitations} ``Compute-environment precision-path dependence''.}
\label{tab:imbalance_baselines_full}
\end{table}

\paragraph{Why ModernBERT collapses under reweighted losses at extreme scarcity.} BERT-for-Patents and ModernBERT respond very differently to the same long-tail objective: focal loss lifts BERT-for-Patents from $0.120$ to $0.483$ (\textbf{a real and substantial gain}), but pushes ModernBERT from $0.314$ down to $0.074$ under the headline bf16+flash-attention-2 setting. The R5 follow-up audit on the revised review-compute environment (last rows of Table~\ref{tab:imbalance_baselines_full}) shows that fp32+eager-attention \emph{partially} recovers focal-loss ModernBERT (to $0.199$) and weighted-BCE ModernBERT (to $0.082$). Two readings of this:
(i) the headline collapse at $0.074$ is partly a precision-path artefact---bf16+flash-attention-2 on this corpus + this loss is numerically pathological---and
(ii) even after fixing precision, focal-loss and weighted-BCE \emph{do not improve} on the plain-BCE baseline of $0.314$. So the qualitative ``reweighted losses do not solve the imbalance on ModernBERT at extreme scarcity'' conclusion is unchanged; the precise micro-F1 number for the failed condition is environment-dependent.

The most plausible mechanism is that aggressive class reweighting amplifies gradient noise rather than rebalancing signal when the training corpus is very small (165 documents) and most labels have $\leq 3$ positive examples. Under focal loss, the $(1-p)^\gamma$ down-weighting of easy examples removes most of the gradient signal that ModernBERT's pretraining-encoded priors provide on the 99\% of label slots that are negative. Under inverse-frequency weighting, the $\texttt{pos\_weight}=100$ multiplier on singleton-label gradients amplifies the variance of an already noisy estimate. BERT-for-Patents starts from a much weaker baseline (0.120 vs.\ 0.314), so its plain-BCE training has less useful signal to disrupt, and any reweighting that increases gradient flow on rare classes is a net win.

This finding---that long-tail loss baselines can hurt rather than help under extreme scarcity---is itself a useful negative result for practitioners. It also makes the synthetic-augmentation rebuttal stronger, not weaker: synthetic data improves both classifiers consistently across all imbalance ratios (Table~\ref{tab:classification_summary}), whereas reweighted losses are classifier- and regime-specific.

\paragraph{Reading the comparison.} The relevant comparison for the ``synthetic vs.\ imbalance handling'' question is the strongest non-augmentation cell in each classifier column. For BERT-for-Patents that is focal loss at $0.483$; synthetic augmentation reaches $0.702$, a $+0.219$ gain on top. For ModernBERT, the strongest non-augmentation cell is plain BCE at $0.314$ (since both reweighted losses degrade it); synthetic augmentation reaches $0.622$, a $+0.308$ gain on top. The synthetic-data gain is therefore not redundant with class-aware loss reweighting on either classifier.

\paragraph{Multi-label-specific long-tail losses (ASL, CB, DB).} Beyond the focal and inverse-frequency baselines reported in Table~\ref{tab:imbalance_baselines_full}, we additionally evaluate three losses explicitly designed for multi-label long-tail classification: Asymmetric Loss (ASL; \citealp{ridnik2021asymmetric}) with $\gamma_{\text{neg}}{=}4$, $\gamma_{\text{pos}}{=}1$, clip${=}0.05$; Class-Balanced BCE \citep{cui2019classbalanced} with effective-number-of-samples reweighting at $\beta{=}0.9999$; and Distribution-Balanced Loss \citep{wu2020distributionbalanced} with NTR boundary shift $\mu{=}0.3$ and focal-style $\gamma{=}2$. All runs use fp32 with eager attention to remove the Flash-Attention-2 + bf16 precision-path confound documented in \S\ref{sec:limitations}. Code, fully runnable on a single L4 GPU, is released as part of the supplementary materials. Table~\ref{tab:long_tail_extension} reports the results at 1:1 and 1:5.

\begin{table}[h]
\centering
\small
\resizebox{\columnwidth}{!}{%
\begin{tabular}{l l c r r}
\toprule
\textbf{Classifier} & \textbf{Loss} & \textbf{Ratio} & \textbf{Micro F1} & \textbf{Macro F1} \\
\midrule
BERT-for-Patents & ASL                       & 1:1 & \textbf{0.353}$\pm$0.086 & 0.241$\pm$0.059 \\
BERT-for-Patents & Class-Balanced BCE        & 1:1 & 0.090$\pm$0.006 & 0.066$\pm$0.002 \\
BERT-for-Patents & Distribution-Balanced     & 1:1 & 0.149$\pm$0.063 & 0.114$\pm$0.052 \\
ModernBERT-base  & ASL                       & 1:1 & 0.114$\pm$0.015 & 0.081$\pm$0.014 \\
ModernBERT-base  & Class-Balanced BCE        & 1:1 & 0.099$\pm$0.009 & 0.061$\pm$0.000 \\
ModernBERT-base  & Distribution-Balanced     & 1:1 & \textbf{0.117}$\pm$0.016 & 0.069$\pm$0.008 \\
\midrule
BERT-for-Patents & ASL                       & 1:5 & \textbf{0.601}$\pm$0.005 & 0.526$\pm$0.013 \\
BERT-for-Patents & Class-Balanced BCE        & 1:5 & 0.574$\pm$0.011 & 0.502$\pm$0.010 \\
BERT-for-Patents & Distribution-Balanced     & 1:5 & 0.451$\pm$0.013 & 0.455$\pm$0.026 \\
ModernBERT-base  & ASL                       & 1:5 & \textbf{0.364}$\pm$0.033 & 0.250$\pm$0.017 \\
ModernBERT-base  & Class-Balanced BCE        & 1:5 & 0.239$\pm$0.016 & 0.181$\pm$0.008 \\
ModernBERT-base  & Distribution-Balanced     & 1:5 & 0.222$\pm$0.019 & 0.154$\pm$0.020 \\
\bottomrule
\end{tabular}%
}
\caption{Multi-label-specific long-tail loss baselines at 1:1 and 1:5 real-only conditions, mean $\pm$ std over 3 seeds. All runs in fp32 with eager attention on a single NVIDIA L4 GPU. \textbf{Bold} marks the strongest non-augmentation cell per classifier/ratio block. ASL dominates the multi-label long-tail family on every cell except ModernBERT at 1:1, where Distribution-Balanced is marginally ahead (within seed noise).}
\label{tab:long_tail_extension}
\end{table}

\paragraph{Environment note.} The absolute F1 magnitudes in Table~\ref{tab:long_tail_extension} are systematically lower than the corresponding cells of Table~\ref{tab:imbalance_baselines_full}---for example, BERT-Pat 1:1 focal in Table~\ref{tab:imbalance_baselines_full} is $0.483$, while recomputed on the unified L4-GPU fp32+eager environment used for Table~\ref{tab:long_tail_extension} it drops to $0.379\pm0.040$. This is the same precision-path / hyperparameter-drift artefact documented in \S\ref{sec:limitations}, not a measurement error in either table. The load-bearing comparison for the multi-label long-tail-loss question is therefore the \emph{within-environment ranking} of Table~\ref{tab:long_tail_extension}: ASL is the strongest multi-label-specific long-tail loss (best non-augmentation BERT-Pat cell at 1:1: $0.353$; best ModernBERT cell at 1:5: $0.364$); CB and DB underperform across the board. Even taking ASL as the ceiling, the synthetic-augmented BERT-for-Patents headline ($0.702$ at 1:1, Table~\ref{tab:classification_summary}) remains $+0.3$ micro F1 above the strongest multi-label long-tail loss. The headline ``synthetic gain is not redundant with class-aware reweighting'' claim therefore holds for the canonical multi-label-specific losses (ASL, CB, DB) as well, not only for focal.

\section{Controlled Comparison with Prior Work}
\label{app:comparison}

Prior work \citep{yousefiramandi2025finetuning} instruction-tuned Llama-3.2-1B-Instruct on 14 vision subcategory labels with LoRA ($r{=}16$, $\alpha{=}16$, dropout 0.05), up to 20 epochs, and DWPI field enrichment, achieving 0.762 F1 on 533 test samples. The present study scales to 64 labels, 2{,}051 test samples, and introduces synthetic augmentation as the central variable. This appendix is \emph{not} a literal reproduction of the cited setup. It compares two SFT configurations on shared splits with multi-seed averaging: a \emph{high-capacity comparison variant} (column key \emph{prev}: LoRA $r{=}64$, DWPI, 5 epochs, batch 1) alongside the \emph{current} configuration used elsewhere in this paper (\emph{curr}: $r{=}32$, no DWPI, 3 epochs, NF4 4-bit, batch 4). Relative to the cited prior work, the high-capacity comparison variant differs in LoRA rank ($r{=}64$ vs.\ cited $r{=}16$) and epoch schedule (5 vs.\ cited up to 20), while matching DWPI use; it is therefore a \emph{higher-rank, fewer-epoch} point along the configuration axis rather than a faithful reproduction. A literal reproduction of the cited $(r{=}16, 20\text{ ep})$ setup is out of scope for this submission and is noted as future work.

\begin{table}[h]
\centering
\footnotesize
\setlength{\tabcolsep}{3pt}
\begin{tabular}{ll rr}
\toprule
\textbf{Config} & \textbf{Aug} & \textbf{Mi-F1} & \textbf{Ma-F1} \\
\midrule
\multicolumn{4}{l}{\textit{14 Vision labels (prior work: 0.762)}} \\
prev ($r$64, DWPI) & No & 0.694\tiny$\pm$0.013 & 0.589\tiny$\pm$0.023 \\
prev ($r$64, DWPI) & Yes & 0.695\tiny$\pm$0.016 & 0.596\tiny$\pm$0.017 \\
curr ($r$32, NF4) & No & 0.678\tiny$\pm$0.008 & 0.457\tiny$\pm$0.021 \\
curr ($r$32, NF4) & Yes & 0.677\tiny$\pm$0.008 & 0.445\tiny$\pm$0.011 \\
\midrule
\multicolumn{4}{l}{\textit{64 labels, original ratio}} \\
prev ($r$64, DWPI) & No & 0.799\tiny$\pm$0.004 & 0.516 \\
prev ($r$64, DWPI) & Yes & \textbf{0.814}\tiny$\pm$0.002 & \textbf{0.584} \\
curr+DWPI ($r$32) & Yes & 0.792\tiny$\pm$0.002 & 0.522 \\
curr ($r$32, NF4) & Yes & 0.790\tiny$\pm$0.005 & 0.497 \\
\midrule
\multicolumn{4}{l}{\textit{64 labels, 1:1 ratio}} \\
prev ($r$64, DWPI) & No & \textbf{0.423}\tiny$\pm$0.002 & 0.157 \\
curr+DWPI ($r$32) & No & 0.323\tiny$\pm$0.005 & 0.078 \\
curr ($r$32, NF4) & No & 0.266\tiny$\pm$0.010 & 0.066 \\
\bottomrule
\end{tabular}
\caption{Controlled comparison of two SFT configurations on shared splits with multi-seed averaging. \emph{prev}: high-capacity comparison variant (LoRA $r{=}64$, DWPI, 5 epochs). \emph{curr}: current ($r{=}32$, NF4 4-bit, 3 epochs). Aug = Qwen2.5-7B paraphrase. The high-capacity comparison variant (0.694) is \emph{not} a literal reproduction of the cited 0.762: it differs from \citet{yousefiramandi2025finetuning} in LoRA rank ($r{=}64$ vs.\ $r{=}16$), epoch count (5 vs.\ up to 20), test-set size (2{,}051 vs.\ 533), and multi-seed averaging. The 0.694 vs.\ 0.762 gap is therefore a combination of configuration changes and data-pipeline differences, not a splits/test-size effect alone.}
\label{tab:comparison}
\end{table}

\paragraph{Configuration sensitivity.} The high-capacity variant (prev\_config) outperforms curr\_config by $+1.6$pp on the 14 vision labels (0.694 vs.\ 0.678) and $+2.4$pp on 64 labels at the original ratio (0.814 vs.\ 0.790). The gap widens at extreme scarcity: $+15.7$pp at 1:1 (0.423 vs.\ 0.266). The larger LoRA rank ($r{=}64$ vs.\ $r{=}32$), more training epochs (5 vs.\ 3), and absence of NF4 quantization provide more model capacity, which matters most when training data is scarce. We do not interpret the 0.694 vs.\ 0.762 gap relative to \citet{yousefiramandi2025finetuning} as a single-cause effect: the high-capacity variant differs from the cited setup in LoRA rank, epoch count, test-set size, and seed averaging simultaneously, so the gap reflects a bundle of configuration and data-pipeline differences rather than splits/test-size alone.

\paragraph{DWPI field enrichment.} Adding proprietary DWPI fields (detailed description, novelty, use, advantage, technology focus) to the current config improves micro F1 by $+0.2$pp at the original ratio and $+5.7$pp at 1:1 (0.266 $\to$ 0.323). We deliberately exclude DWPI from our main experiments to ensure reproducibility with publicly available patent data.

\paragraph{Augmentation interaction.} Synthetic augmentation (Qwen2.5-7B paraphrase) helps the SFT classifier modestly at the original ratio ($+1.5$pp for prev\_config, $+2.1$pp for curr\_config) but is ineffective or slightly harmful at 1:1. This is consistent with the main-paper finding that instruction-tuned models benefit less from synthetic augmentation than discriminative BERT-family classifiers.

\section{Embedding Metrics Across All Ratios}
\label{app:embedding_metrics}

Table~\ref{tab:embedding_all_ratios} shows MMD (PatentSBERTa) across all imbalance ratios.

\begin{table*}[t]
\centering
\footnotesize
\setlength{\tabcolsep}{4pt}
\begin{tabular}{l rrrr rrrr}
\toprule
& \multicolumn{4}{c}{\textbf{Full Synthetic (MMD}$\downarrow$\textbf{)}} & \multicolumn{4}{c}{\textbf{Paraphrase (MMD}$\downarrow$\textbf{)}} \\
\cmidrule(lr){2-5} \cmidrule(lr){6-9}
\textbf{Model} & \textbf{1:1} & \textbf{1:5} & \textbf{1:10} & \textbf{Orig} & \textbf{1:1} & \textbf{1:5} & \textbf{1:10} & \textbf{Orig} \\
\midrule
Qwen2.5-7B & 0.070 & 0.025 & 0.017 & 0.005 & 0.023 & 0.008 & 0.006 & 0.003 \\
Qwen3-4B & 0.083 & 0.031 & 0.021 & 0.007 & 0.030 & 0.010 & 0.007 & 0.003 \\
Llama-3.1-8B & 0.060 & 0.022 & 0.014 & 0.005 & 0.037 & 0.012 & 0.008 & 0.003 \\
Phi-4-mini & 0.069 & 0.024 & 0.016 & 0.005 & 0.026 & 0.008 & 0.006 & 0.003 \\
Gemma-3-12b & 0.076 & 0.028 & 0.019 & 0.006 & 0.033 & 0.011 & 0.007 & 0.003 \\
Mistral-7B & 0.066 & 0.023 & 0.016 & 0.005 & 0.020 & 0.007 & 0.005 & 0.002 \\
\bottomrule
\end{tabular}
\caption{MMD (PatentSBERTa) across all imbalance ratios and strategies. MMD decreases as real training data increases, consistent with better distribution coverage. Paraphrasing achieves lower MMD across all conditions.}
\label{tab:embedding_all_ratios}
\end{table*}

\section{LLM-as-Judge Quality Assessment}
\label{app:llm_judge}

We evaluate synthetic patent quality using three LLM judges (GPT-5-4, Claude Opus 4.6, Gemini 2.5 Pro) served via Databricks AI Gateway (hence the \texttt{databricks-} prefix in model identifiers), scoring 16K samples on technical plausibility, label consistency, and novelty (1--5 scale). Each judge receives the following prompt at temperature 0.0:

\begin{quote}\small
\textit{You are an expert patent examiner evaluating the quality of a synthetic patent document. Rate the following synthetic patent on three dimensions using a 1--5 scale.}

\textit{Patent to Evaluate: Title: [...] Abstract: [...] First Claim: [...] Assigned Labels: [...]}

\textit{Rating Criteria:}
\textit{1.\ \textbf{Technical Plausibility (1--5)}: Does this read like a real patent? (1: Clearly fake, nonsensical; 3: Somewhat plausible but with noticeable issues; 5: Indistinguishable from a real patent)}
\textit{2.\ \textbf{Label Consistency (1--5)}: Do the assigned labels match the content? (1: Labels completely wrong; 3: Some match; 5: All labels perfectly match)}
\textit{3.\ \textbf{Novelty (1--5)}: Does this describe a unique invention? (1: Generic/template-like; 3: Some specific details but formulaic; 5: Specific, novel technical approach)}

\textit{Respond with ONLY a JSON object: \{``technical\_plausibility'': $\langle$int$\rangle$, ``label\_consistency'': $\langle$int$\rangle$, ``novelty'': $\langle$int$\rangle$\}}
\end{quote}

\begin{figure*}[t]
\centering
\includegraphics[width=\textwidth]{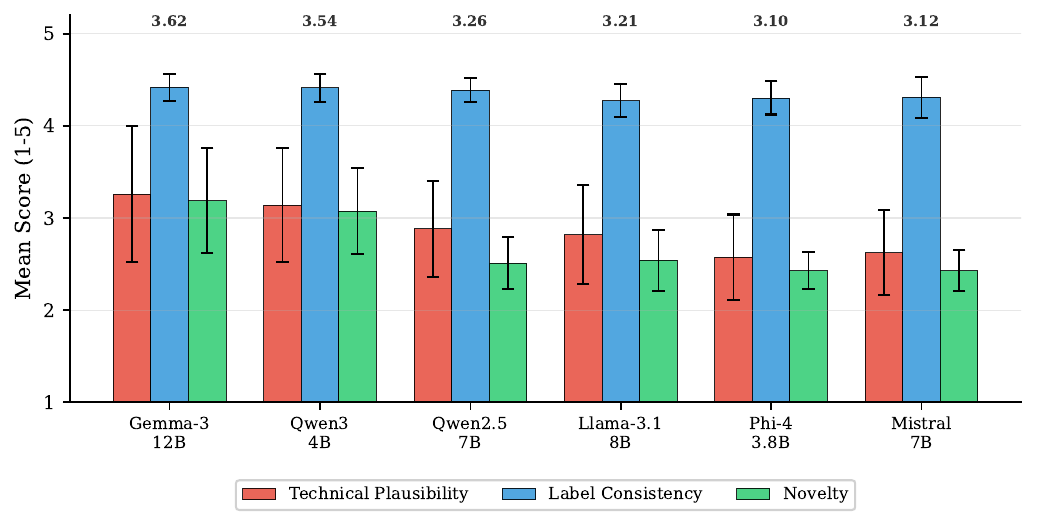}
\caption{Mean LLM judge quality scores (1--5) by generator model, averaged across three judges (GPT-5-4, Claude Opus 4.6, Gemini 2.5 Pro). Error bars show inter-judge standard deviation. Label consistency is uniformly high (${\sim}$4.2--4.5) across all generators, while technical plausibility and novelty vary more (2.1--4.1). Gemma-3-12b and Qwen3-4B receive the highest overall scores (3.62 and 3.54), though these weakly predict classification utility ($r{=}0.24$). Numbers above each cluster show the overall average across all three dimensions.}
\label{fig:judge_boxplot}
\end{figure*}

\begin{figure*}[t]
\centering
\begin{subfigure}[t]{0.32\textwidth}
\includegraphics[width=\textwidth]{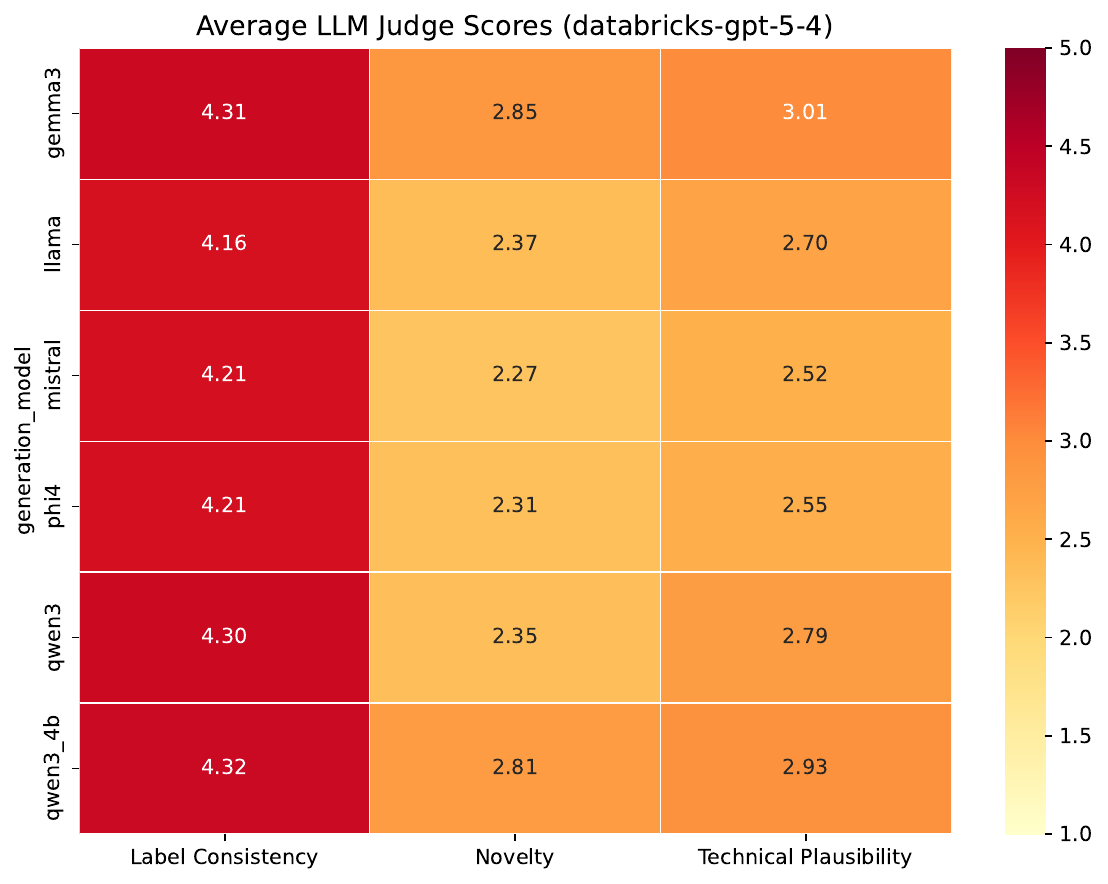}
\caption{GPT-5-4}
\end{subfigure}
\hfill
\begin{subfigure}[t]{0.32\textwidth}
\includegraphics[width=\textwidth]{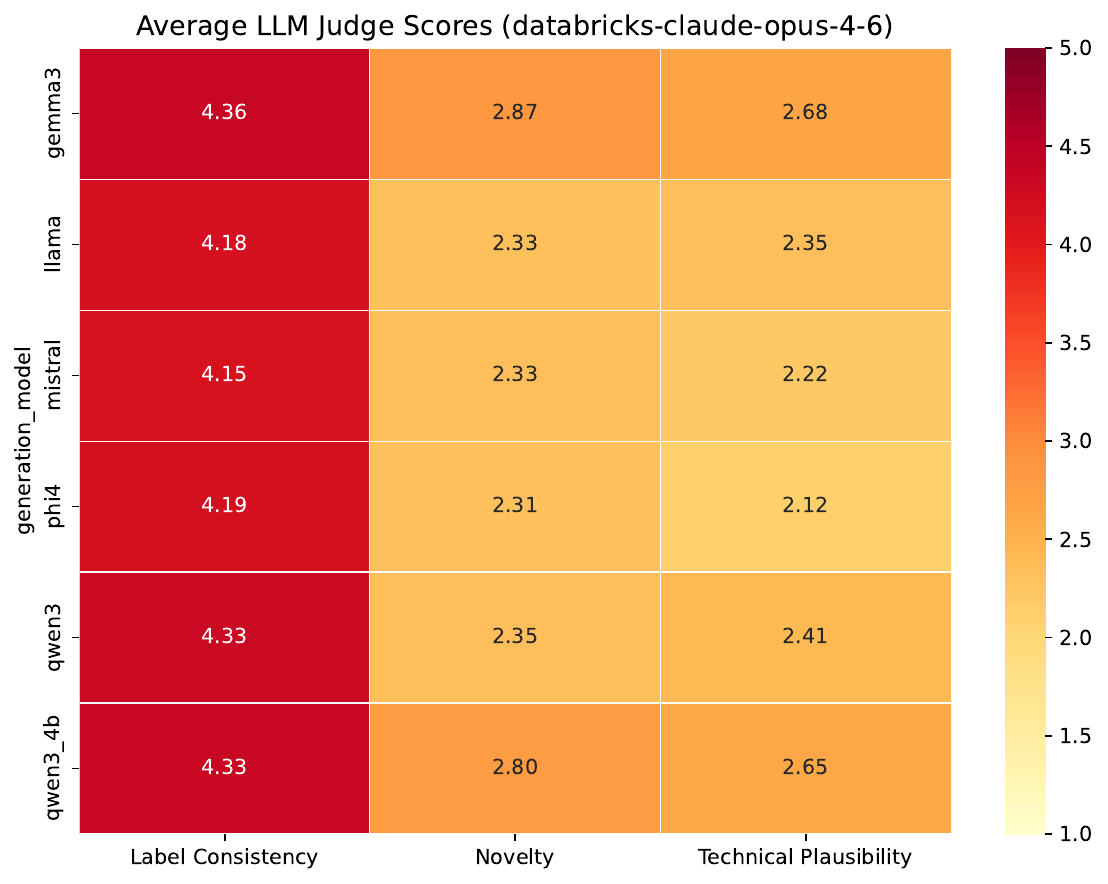}
\caption{Claude Opus 4.6}
\end{subfigure}
\hfill
\begin{subfigure}[t]{0.32\textwidth}
\includegraphics[width=\textwidth]{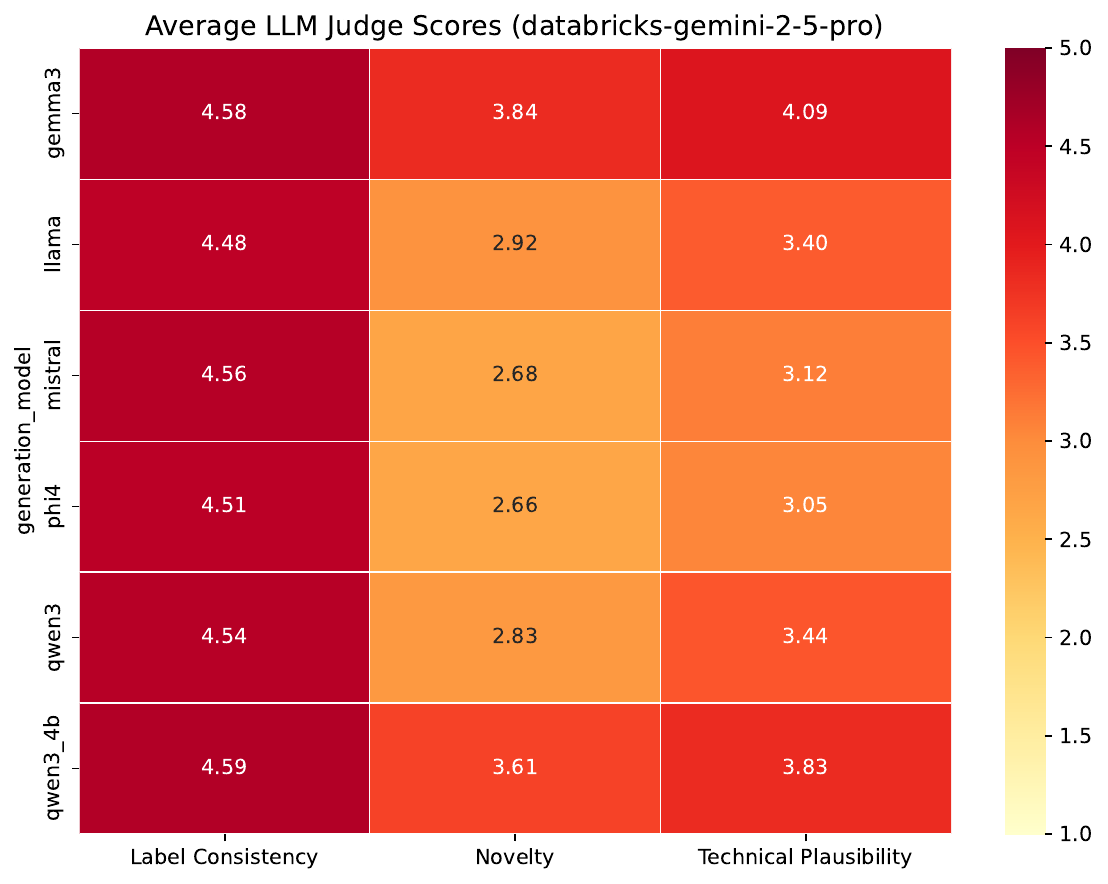}
\caption{Gemini 2.5 Pro}
\end{subfigure}
\caption{Mean LLM judge scores by generator model and quality dimension. Label consistency is uniformly high across all generators (4.2--4.6), while technical plausibility varies more (2.1--4.0). Gemma-3-12b scores highest on all three dimensions across all judges. The three judges agree on relative generator rankings despite differing in absolute scale (Gemini is the most lenient).}
\label{fig:judge_heatmap}
\end{figure*}

Figure~\ref{fig:judge_boxplot} shows the mean scores per generator and quality dimension, averaged across all three judges (error bars show inter-judge standard deviation). Figure~\ref{fig:judge_heatmap} shows the per-judge breakdown.

\section{Ensemble Augmentation}
\label{app:ensemble}

\begin{figure}[t]
\centering
\includegraphics[width=\columnwidth]{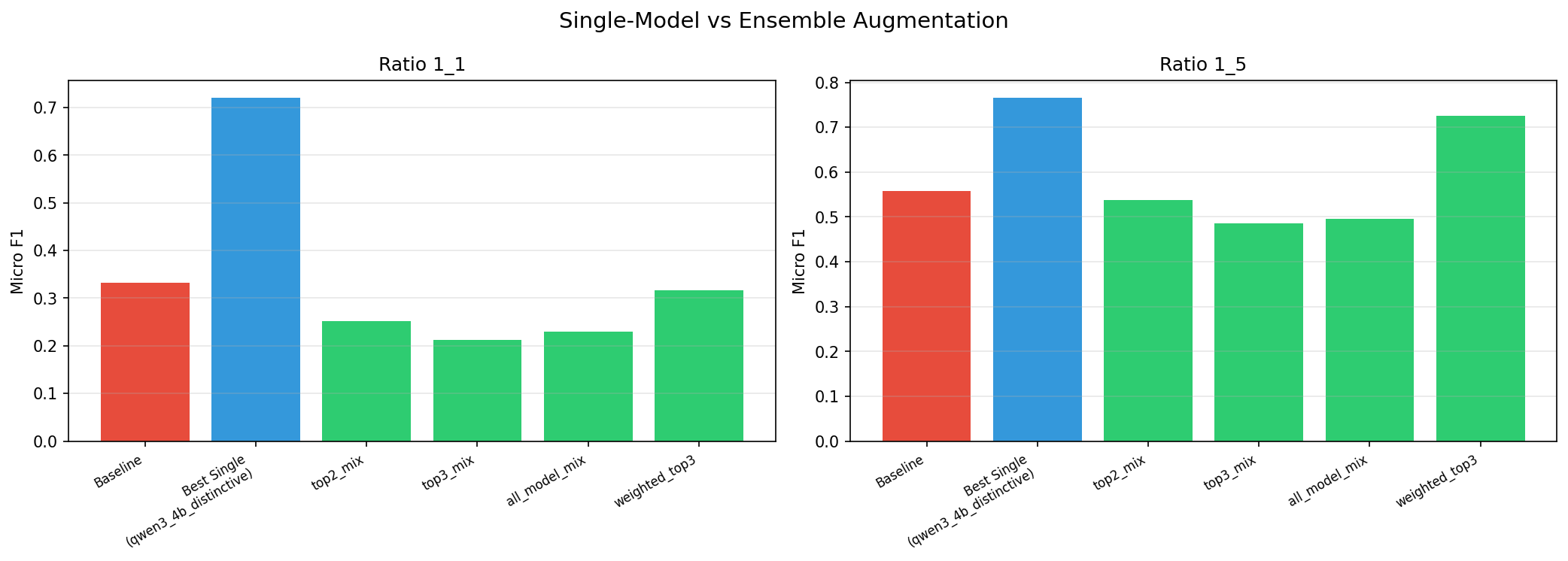}
\caption{Single-model vs.\ ensemble augmentation. At both 1:1 and 1:5 ratios, the single best generator (green) outperforms all ensemble combinations (red). The weighted top-3 ensemble approaches single-model performance at 1:5 but does not exceed it.}
\label{fig:ensemble}
\end{figure}

We test whether combining synthetic data from multiple generators improves over single-model augmentation. Figure~\ref{fig:ensemble} compares single-model (best generator) against ensemble strategies (top-2, top-3, all-6, weighted top-3).

\section{Per-Label Augmentation Heatmap and Bar Chart}
\label{app:per_label_heatmap}

\begin{figure}[t]
\centering
\includegraphics[width=\columnwidth]{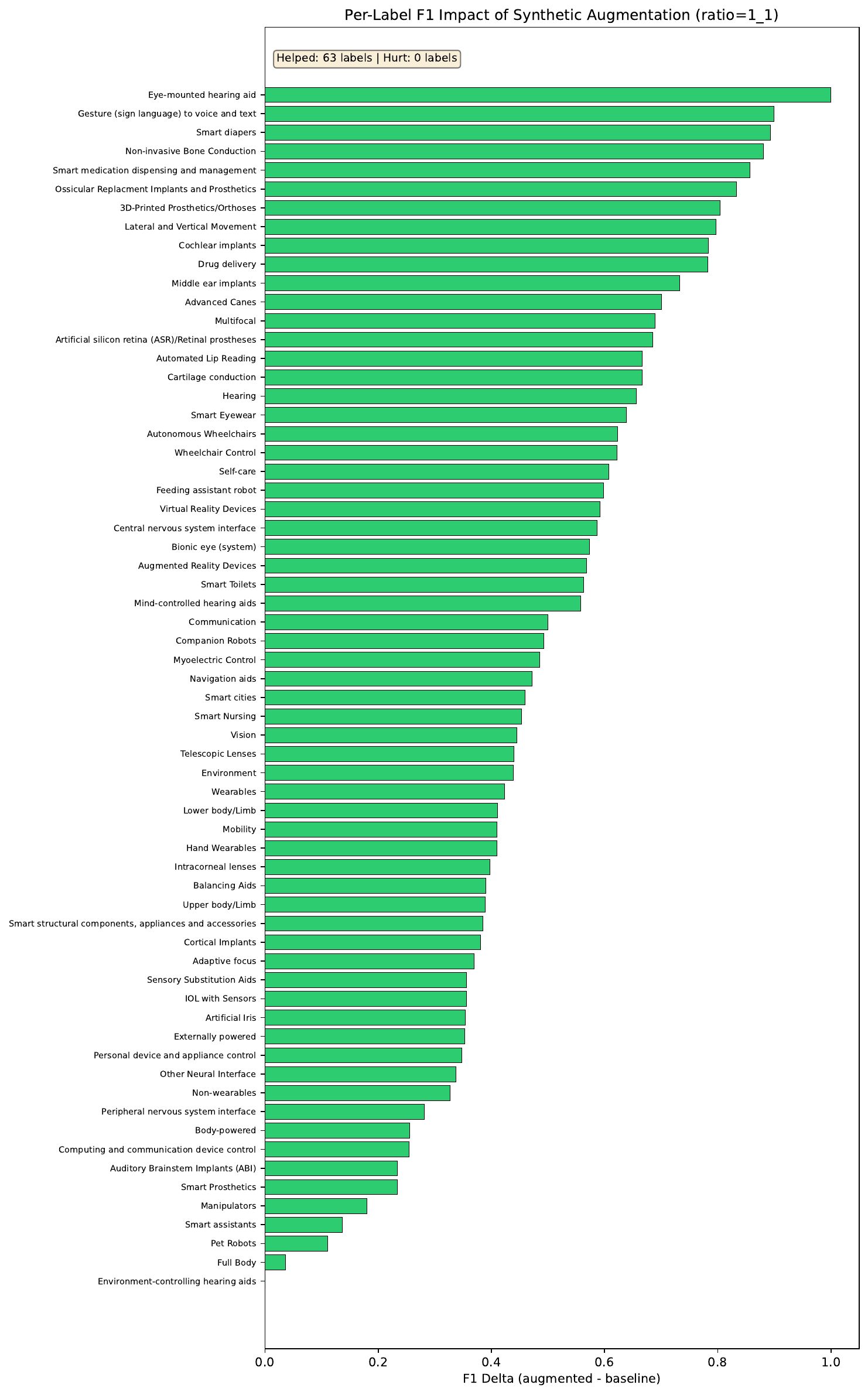}
\caption{Per-label F1 improvement at the 1:1 ratio (63 non-zero-baseline labels). No label is hurt at 1:1; top gainers reach $+0.83$--$+0.99$. Realized per-label positive counts (not target ratios) are used for ordering.}
\label{fig:per_label_bar_app}
\end{figure}

\begin{figure*}[t]
\centering
\includegraphics[width=\textwidth]{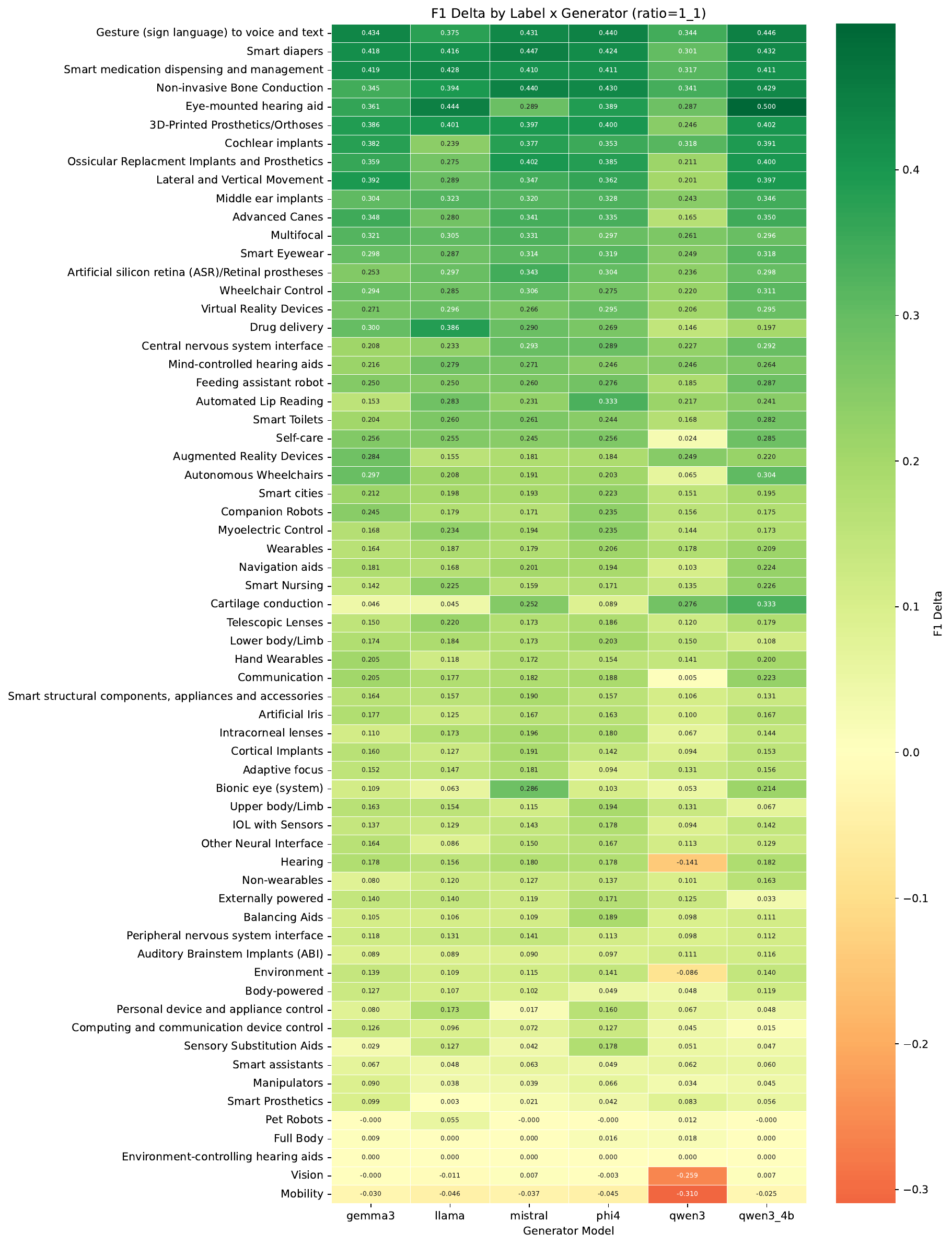}
\caption{Per-label F1 delta (augmented $-$ baseline) at the 1:1 ratio, broken down by generator model. Rows are labels (sorted by average delta); columns are generators. Most labels improve across all generators (green), but a few (Full Body, Environment-controlling hearing aids) show near-zero gains regardless of generator.}
\label{fig:per_label_heatmap}
\end{figure*}

Figure~\ref{fig:per_label_bar_app} shows the full per-label F1 improvement at the 1:1 ratio (seed-averaged across generators, full synthetic) for the 63 labels with non-zero baseline F1; the one remaining label (\emph{Smart Diapers}) has a zero baseline and is reported in the main text ($0\to 0.813$). Figure~\ref{fig:per_label_heatmap} further breaks the per-label delta down by generator.

\section{Cross-Model Augmentation Benefit}
\label{app:cross_model}

\begin{figure*}[t]
\centering
\includegraphics[width=\textwidth]{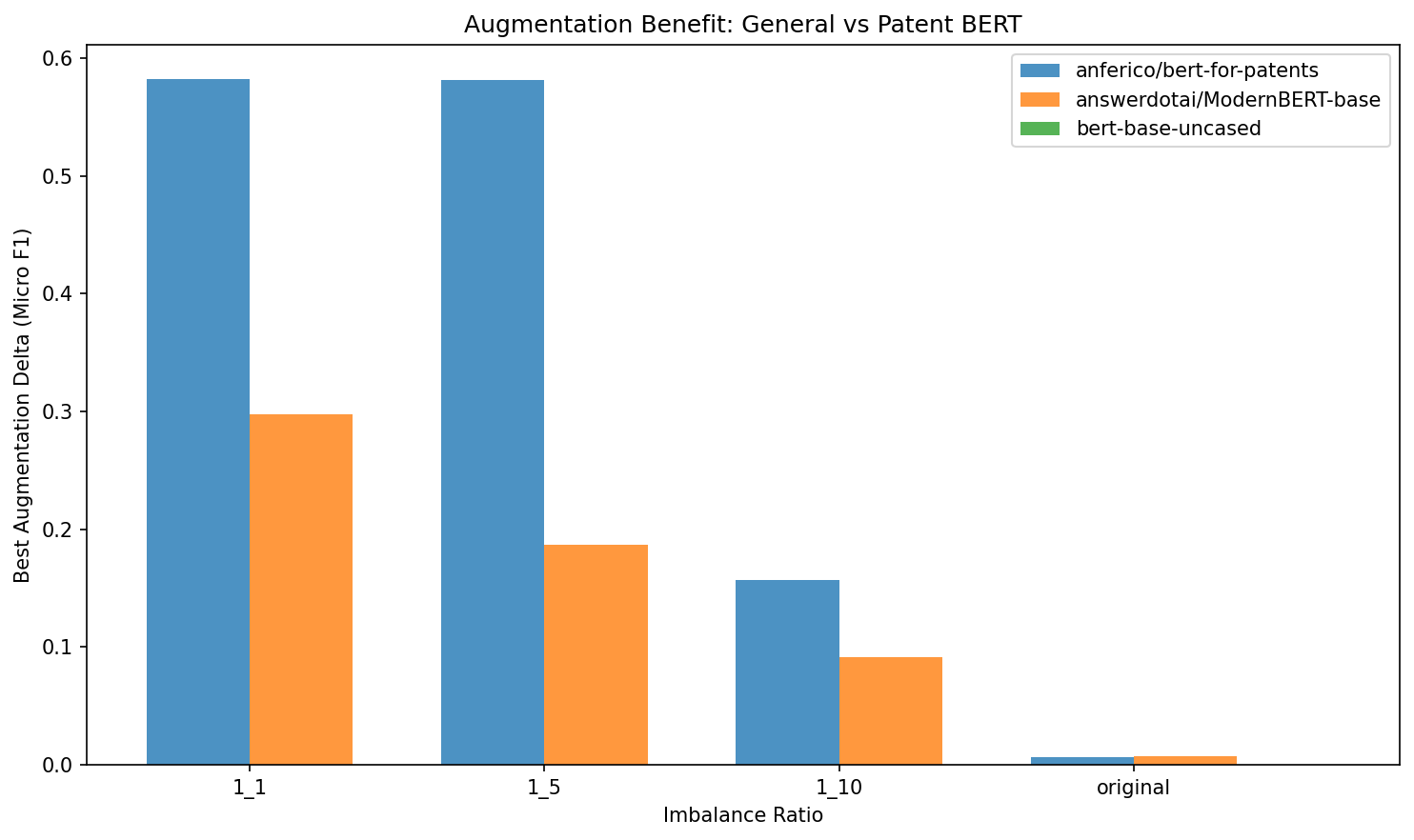}
\caption{Best augmentation delta (micro F1) by classifier model and imbalance ratio. BERT-for-Patents shows the largest absolute gains at low-data ratios, while ModernBERT starts from a higher baseline. SFT benefits least from augmentation across all ratios.}
\label{fig:cross_model_delta}
\end{figure*}
\FloatBarrier

Figure~\ref{fig:cross_model_delta} compares the augmentation benefit across different classifier architectures, showing how the best-case F1 improvement varies by imbalance ratio and classifier model. Table~\ref{tab:bert_detailed} gives the corresponding per-generator detail for BERT-for-Patents; Table~\ref{tab:modernbert_detailed} in Appendix~\ref{app:modernbert_detailed} gives the same detail for ModernBERT.

\begin{table}[t]
\centering
\small
\begin{tabular}{l rr}
\toprule
\textbf{Model / Strategy} & \textbf{1:1} & \textbf{1:5} \\
\midrule
\textit{Baseline (real only)} & \textit{0.120} & \textit{0.182} \\
\midrule
\multicolumn{3}{l}{\textbf{Full Synthetic}} \\
\quad Qwen3-4B (distinctive) & \textbf{0.702} & 0.757 \\
\quad Qwen3-4B & 0.693 & 0.754 \\
\quad Mistral-7B & 0.679 & 0.738 \\
\quad Phi-4-mini & 0.676 & \textbf{0.758} \\
\quad Gemma-3-12b & 0.667 & 0.738 \\
\quad Qwen2.5-7B & 0.663 & 0.748 \\
\quad Llama-3.1-8B & 0.631 & 0.733 \\
\midrule
\multicolumn{3}{l}{\textbf{Paraphrase}} \\
\quad Qwen2.5-7B & 0.000 & \textbf{0.763} \\
\quad Gemma-3-12b & 0.000 & 0.759 \\
\quad Llama-3.1-8B & 0.000 & 0.759 \\
\quad Qwen3-4B (distinctive) & 0.000 & 0.758 \\
\quad Qwen3-4B & 0.000 & 0.757 \\
\quad Phi-4-mini & 0.000 & 0.756 \\
\quad Mistral-7B & 0.000 & 0.754 \\
\bottomrule
\end{tabular}
\caption{BERT-for-Patents micro F1 for all generator--strategy combinations at the 1:1 and 1:5 ratios. At 1:1, only full synthesis reaches non-trivial F1 (paraphrase yields zero F1 across all generators due to insufficient volume---${\sim}400$ paraphrases from 165 source documents). At 1:5, both strategies are competitive with a generator spread of only 0.030.}
\label{tab:bert_detailed}
\end{table}

\section{Curriculum Learning Strategies}
\label{app:curriculum}

\begin{figure*}[t]
\centering
\includegraphics[width=\textwidth]{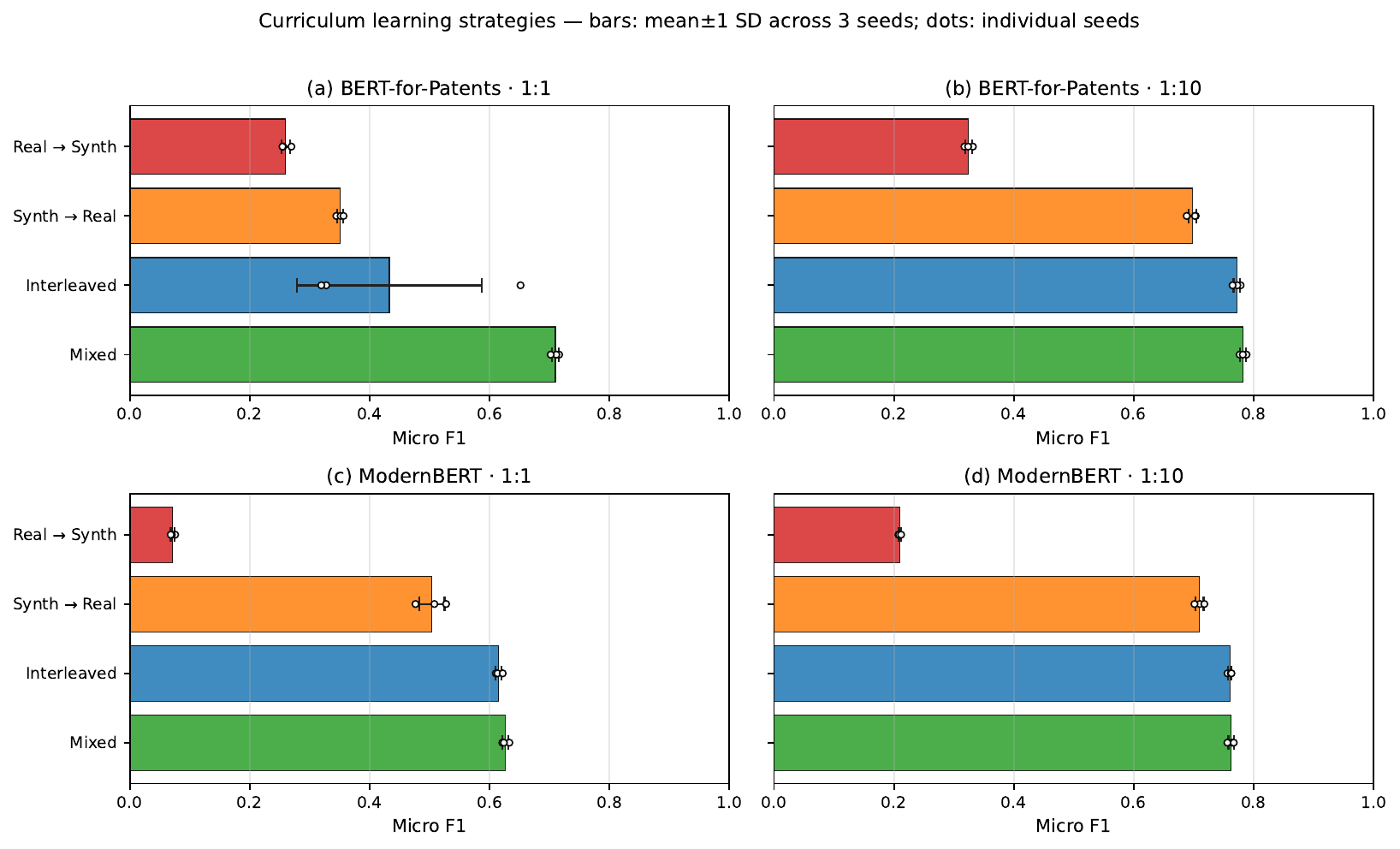}
\caption{Curriculum learning strategies (micro F1) for BERT-for-Patents (top) and ModernBERT (bottom) at 1:1 (left) and 1:10 (right) ratios. Mixed training dominates for both architectures. BERT interleaved shows extreme variance at 1:1 ($\sigma{>}0.17$), while ModernBERT interleaved is stable ($\sigma{<}0.01$). Real$\to$Synth is catastrophic for both, especially ModernBERT (0.06 at 1:1). Bars show mean $\pm$1 SD across 3 seeds; black dots show the three individual seed results so that the spread underlying each mean is directly visible.}
\label{fig:curriculum}
\end{figure*}

Figure~\ref{fig:curriculum} shows the full curriculum learning comparison for both classifiers.

\section{TSTR Substitution Details}
\label{app:tstr}

\begin{figure*}[t]
\centering
\includegraphics[width=\textwidth]{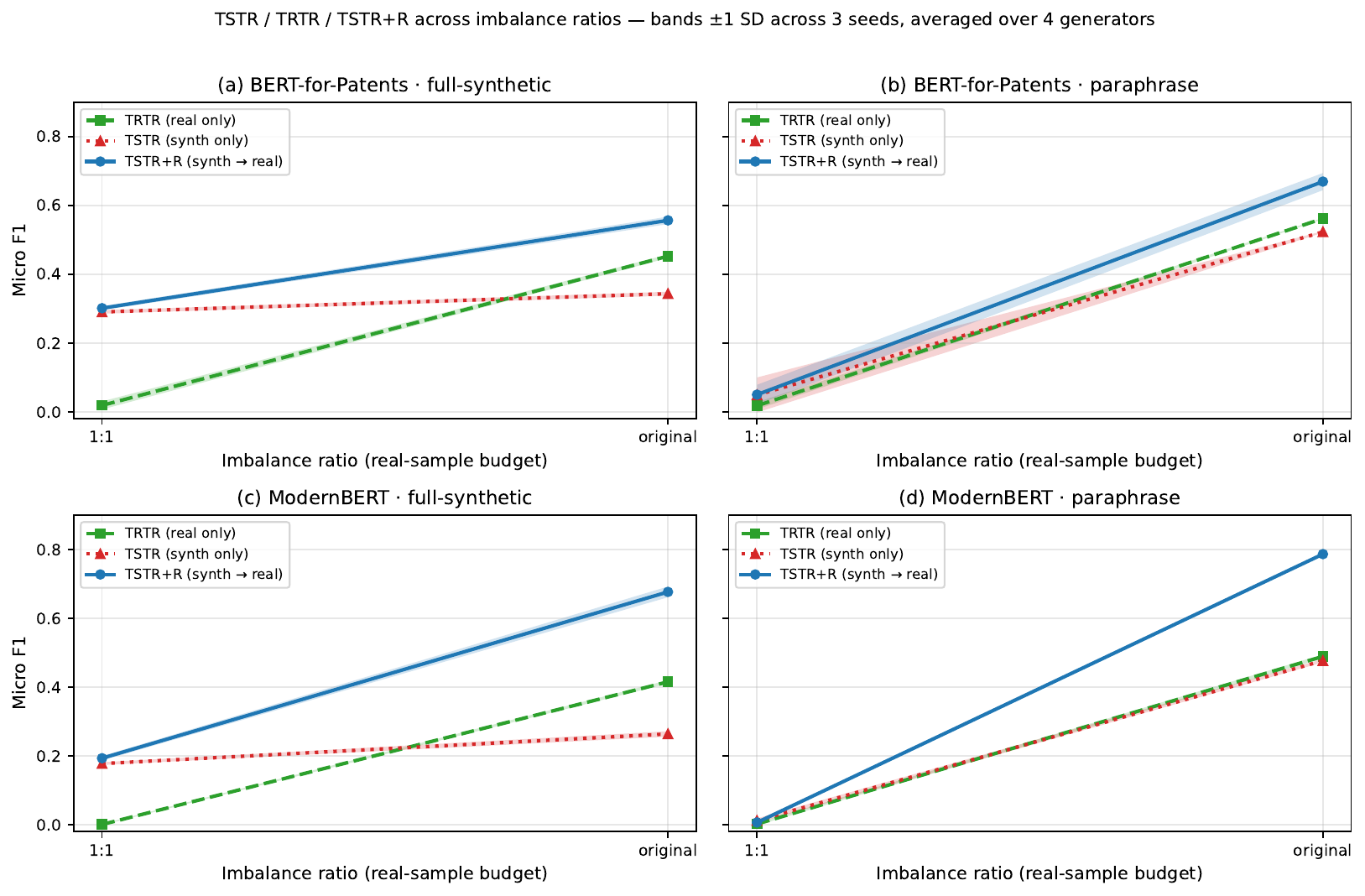}
\caption{TSTR (red, synth-only), TRTR (green, real-only), and TSTR+R (blue, synth pretrain $\rightarrow$ real finetune) across imbalance ratios for BERT-for-Patents (top) and ModernBERT (bottom) under full-synthetic (left) and paraphrase (right) strategies. Each line averages 4 generators (Qwen3-4B, Phi-4, Llama-3.1-8B, Gemma-3); shaded bands show $\pm$1 SD across 3 seeds.}
\label{fig:tstr}
\end{figure*}

Figure~\ref{fig:tstr} shows the full TSTR / TRTR / TSTR+R comparison across imbalance ratios for both classifiers and both strategies. The blue curve (TSTR+R: synthetic pretrain, real finetune) dominates both the synth-only (red) and the real-only baseline (green) at the original ratio under paraphrase, reaching 0.787 for ModernBERT vs.\ a real-only budget of 0.489 trained on the same 1,000-sample subset---evidence that synthetic pretraining leverages scale even when real data is abundant. At the 1:1 ratio paraphrase yields near-zero F1 for all three regimes under BERT-for-Patents, reproducing the degeneracy discussed in \S\ref{sec:discussion}.

\section{CQF Threshold Sweep}
\label{app:cqf}

\begin{figure*}[t]
\centering
\includegraphics[width=\textwidth]{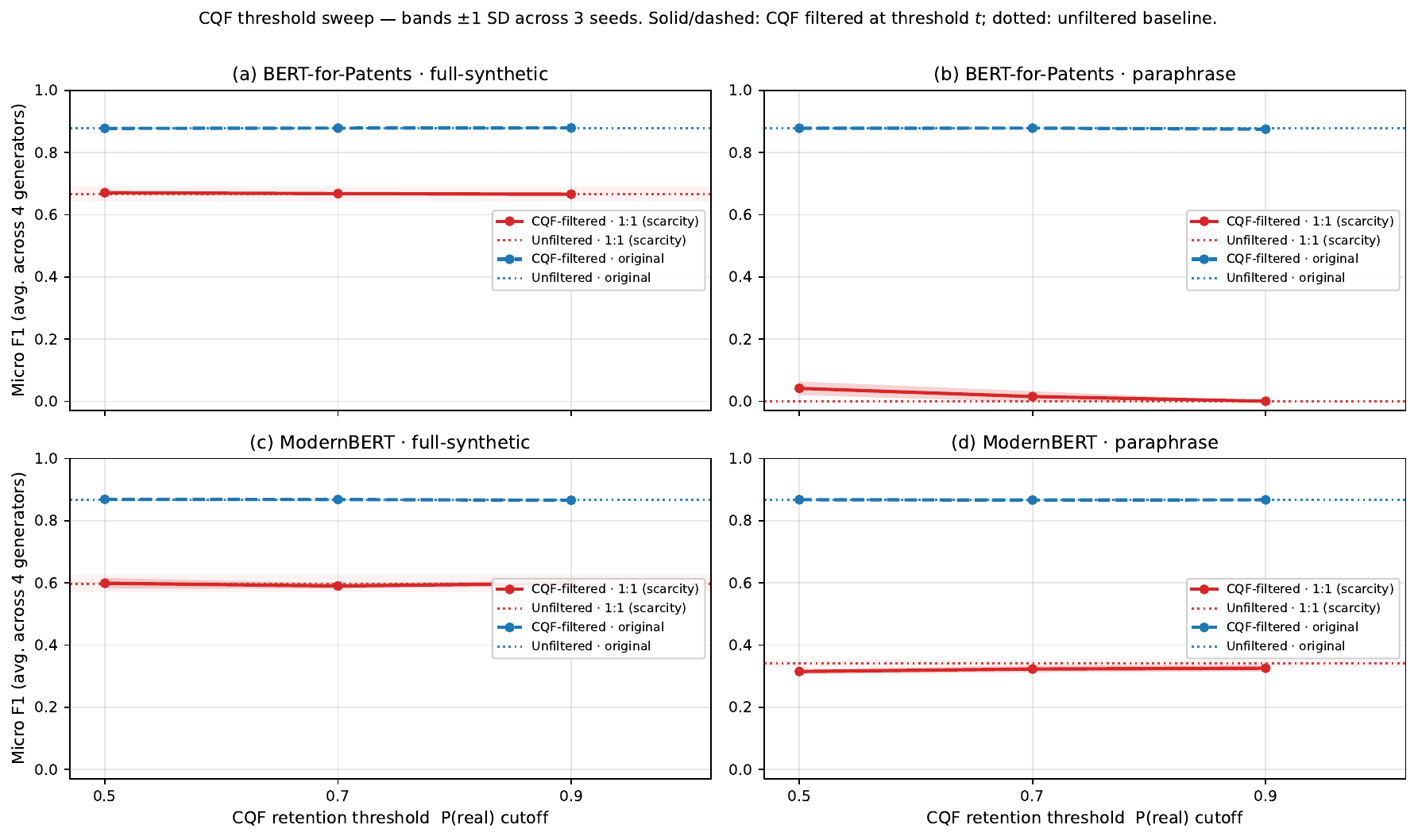}
\caption{CQF retention-threshold sweep (P(real) cutoff $\in \{0.5, 0.7, 0.9\}$) for BERT-for-Patents (top) and ModernBERT (bottom), under full-synthetic (left) and paraphrase (right) strategies. Solid red: 1:1 ratio, dashed blue: original ratio. Dotted lines give the unfiltered augmentation baseline at the matching ratio. Filtered curves average across 4 generators (Qwen3-4B, Phi-4, Llama-3.1-8B, Gemma-3); shaded bands show $\pm$1 SD across 3 seeds.}
\label{fig:cqf}
\end{figure*}

Figure~\ref{fig:cqf} shows how classifier-based quality filtering (CQF) performance tracks the retention threshold $t$. For every classifier/strategy pair, the CQF-filtered curve sits at or below the unfiltered baseline (dotted horizontal) across the full threshold range $t\in\{0.5, 0.7, 0.9\}$, confirming that CQF is counterproductive or neutral rather than beneficial: the upstream rule-based filters (length, deduplication, leakage detection) already capture the available quality signal, and further classifier-based pruning either removes useful samples (at 1:1 scarcity) or has no effect (at the original ratio where augmentation itself is noise).

\section{Detailed Retrieval Analysis}
\label{app:retrieval_detailed}

This appendix expands the retrieval paragraph in \S\ref{sec:results_retrieval_brief}. We report three related analyses: (i) retrieval with synthetic \emph{added} to the corpus (as in the main paper), (ii) a fixed-corpus-size control that separates genuine signal-harm from corpus dilution, and (iii) a training-only embedding fine-tuning analysis that avoids the corpus question entirely.

\subsection{Corpus-Augmented Retrieval}

Table~\ref{tab:retrieval_app} reports the main-paper numbers. Adding 22K--28K synthetic documents to a 13{,}667-document real corpus degrades nDCG@10 by $-25$ to $-37\%$ and collapses Recall@10 by an order of magnitude for full synthesis. Paraphrase-augmented corpora degrade less but are still below real-only.

\begin{table}[h]
\centering
\small
\resizebox{\columnwidth}{!}{%
\begin{tabular}{ll rrr}
\toprule
\textbf{Corpus} & \textbf{Strategy} & \textbf{nDCG@10} & \textbf{MRR} & \textbf{R@10} \\
\midrule
\multicolumn{5}{l}{\textit{PatentSBERTa embeddings, 1:1 ratio}} \\
Real only & -- & \textbf{0.525} & \textbf{0.801} & \textbf{0.215} \\
Real+Synth & Full synth & 0.360 & 0.631 & 0.011 \\
Real+Synth & Paraphrase & 0.396 & 0.720 & 0.134 \\
\midrule
\multicolumn{5}{l}{\textit{Nemotron embeddings, 1:1 ratio}} \\
Real only & -- & \textbf{0.648} & \textbf{0.880} & \textbf{0.268} \\
Real+Synth & Full synth & 0.416 & 0.684 & 0.013 \\
Real+Synth & Paraphrase & 0.474 & 0.772 & 0.155 \\
\bottomrule
\end{tabular}%
}
\caption{Corpus-augmented retrieval (same numbers as the main-paper summary in \S\ref{sec:results_retrieval_brief}).}
\label{tab:retrieval_app}
\end{table}

\subsection{Fixed-Corpus-Size Control}
\label{app:retrieval_fixed_size}

The corpus-augmented setup inflates corpus size by 40--50\%, which is itself a retrieval-harming factor (more distractors per query). We construct a fixed-size control in which the final retrieval corpus has the same number of documents as the real-only corpus: for each synthetic document added, one random real document is held out of the index (real documents remain in the query pool and in the ground-truth relevance labels for other corpus docs). \emph{Identification caveat.} Because the held-out real documents may themselves be relevant targets for queries that remain in the pool, this control mixes at least three effects---corpus dilution, synthetic--real distributional mismatch, and asymmetric removal of relevant real targets from the index. We therefore treat the resulting numbers as a weaker robustness check rather than a clean dilution-vs-signal decomposition. A cleaner control (e.g., evaluating only on queries whose relevant real neighbours remain in the index, or constructing matched candidate pools that do not remove relevant real documents asymmetrically) is left to future work.

\begin{table}[h]
\centering
\small
\resizebox{\columnwidth}{!}{%
\begin{tabular}{ll rrr}
\toprule
\textbf{Corpus (fixed size)} & \textbf{Strategy} & \textbf{nDCG@10} & \textbf{MRR} & \textbf{R@10} \\
\midrule
Real only (ref) & -- & 0.525 & 0.801 & 0.215 \\
Synth-subst.\ 50\% & Full synth & 0.451 & 0.738 & 0.152 \\
Synth-subst.\ 50\% & Paraphrase & 0.488 & 0.771 & 0.181 \\
\bottomrule
\end{tabular}%
}
\caption{Fixed-corpus-size substitution (1:1, PatentSBERTa). With corpus size held fixed, synthetic-augmented retrieval is still below real-only, and the gap is smaller than in the corpus-augmented view. The remaining drop is consistent with---but does not cleanly isolate---genuine synthetic--real distributional mismatch (see identification caveat in the setup paragraph).}
\label{tab:retrieval_fixed}
\end{table}

\paragraph{Interpretation.} With corpus size controlled, full synthesis still costs $-0.074$ nDCG@10 ($-14\%$) and paraphrase $-0.037$ ($-7\%$) relative to real-only, vs.\ $-0.165$ ($-31\%$) and $-0.129$ ($-25\%$) in the corpus-augmented view. The corpus-augmented gap shrinks substantially when corpus size is held fixed, but---because the fixed-size control also asymmetrically removes relevant real targets from the index---a precise dilution-vs-signal decomposition is not identified by this control. We therefore down-weight the retrieval claim in the main body (\S\ref{sec:results_retrieval_brief}) and frame the headline as ``classification utility does not imply retrieval utility'' rather than a quantitative magnitude claim.

\subsection{Training-Only Embedding Fine-Tuning}
\label{app:retrieval_train_only}

To test whether synthetic data can improve the \emph{retrieval model itself} (rather than merely pollute the corpus), we fine-tune Qwen3-Embedding-0.6B on co-label taxonomy pairs derived from different data sources and evaluate on the original real-only corpus.

\begin{table}[h]
\centering
\small
\resizebox{\columnwidth}{!}{%
\begin{tabular}{ll rrr}
\toprule
\textbf{Condition} & \textbf{Ratio} & \textbf{nDCG@10} & \textbf{MRR} & \textbf{R@10} \\
\midrule
Baseline (no ft) & -- & 0.783 & 0.880 & 0.027 \\
Real-only & original & \textbf{0.868} & \textbf{0.908} & \textbf{0.034} \\
Synth-only & original & 0.756 & 0.857 & 0.026 \\
Synth-only & 1:1 & 0.706 & 0.817 & 0.023 \\
Blended & original & \textbf{0.869} & 0.904 & \textbf{0.034} \\
Blended & 1:1 & 0.776 & 0.860 & 0.028 \\
\bottomrule
\end{tabular}%
}
\caption{Embedding fine-tuning with Qwen3-Embedding-0.6B on co-label taxonomy pairs (${\sim}100$K pairs each). Evaluation is on the real validation corpus. Real-only and blended (original) fine-tuning improve nDCG@10 by $+11\%$. Synthetic-only fine-tuning \emph{degrades} retrieval ($-3.4\%$ to $-9.7\%$), confirming that the synthetic--real distributional gap corrupts learned representations---independent of corpus size.}
\label{tab:embedding_ft_app}
\end{table}

The training-only setup isolates signal-harm from corpus-size effects: corpus is fixed, only the embedding parameters change. The persistence of negative effects here is the cleanest evidence that synthetic patent pairs are a worse training signal than real ones.

\subsection{Discussion}

Our reading of these three analyses is in \S\ref{sec:discussion_retrieval}: because three of four prompt families generate off-genre technical text (FAQ, Summary, Comparative Analysis), the synthetic corpus both looks less patent-like on the surface and, once integrated into an embedding space, pulls the representation of real patents off the manifold that retrieval queries depend on. A pure-patent-style variant would be the cleanest test of this hypothesis and is our top retrieval-side follow-up.

\section{Embedding Fine-Tuning for Retrieval}
\label{app:embedding_ft}

\begin{figure}[t]
\centering
\includegraphics[width=\columnwidth]{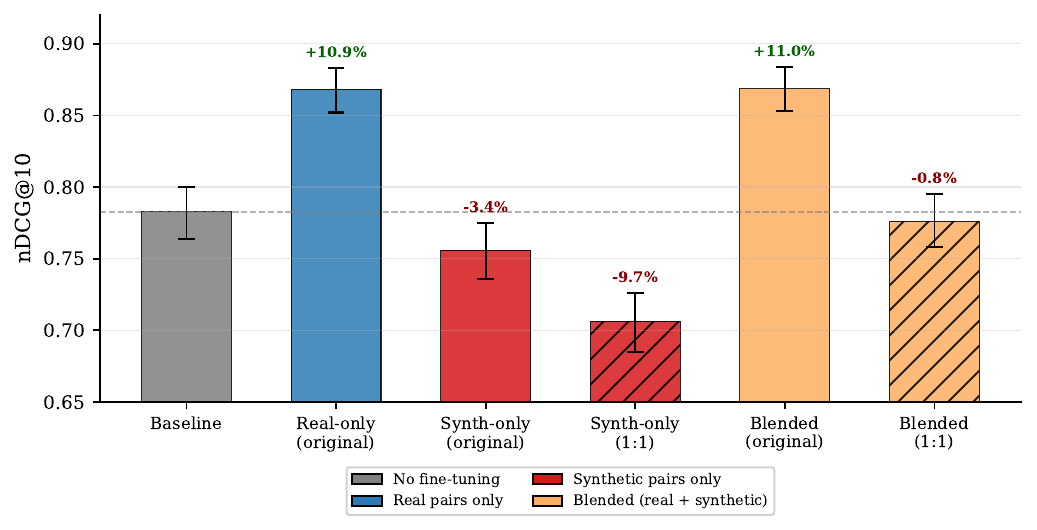}
\caption{Embedding fine-tuning impact on retrieval (nDCG@10). Fine-tuning Qwen3-Embedding-0.6B on real patent pairs improves retrieval by +10.9\%. Synthetic-only fine-tuning \textit{degrades} performance ($-$3.4\% to $-$9.7\%), while blended fine-tuning at the original ratio matches real-only. Hatched bars indicate the 1:1 (extreme scarcity) ratio. Error bars show 95\% bootstrap CIs.}
\label{fig:embedding_ft}
\end{figure}

Figure~\ref{fig:embedding_ft} shows the retrieval performance (nDCG@10) across all embedding fine-tuning conditions.

\section{Label Taxonomy}
\label{app:labels}

The 64 labels comprise 6 domain-level categories: Hearing, Vision, Mobility, Communication, Environment, and Self-care. The 58 subcategory labels include: Smart Diapers, Externally Powered prosthetics, Hand Wearables, Companion Robots, Cochlear Implants, Auditory Brainstem Implants, Non-invasive Bone Conduction, Wheelchair Control, Smart Prosthetics, 3D-Printed Prosthetics, Autonomous Wheelchairs, Bionic Eye systems, Smart Eyewear, Virtual Reality Devices, Augmented Reality Devices, Cortical Implants, Artificial Silicon Retina, Navigation Aids, Smart Medication Dispensing, and 39 additional subcategories.

\section{ModernBERT Detailed Results}
\label{app:modernbert_detailed}

Table~\ref{tab:modernbert_detailed} presents the full ModernBERT-base results for all model--strategy combinations at the 1:1 and 1:5 ratios, complementing the BERT-for-Patents detailed results in Table~\ref{tab:bert_detailed}. ModernBERT shows a narrower generator spread than BERT-for-Patents (0.064 vs.\ 0.071 at 1:1 for full synthesis), and paraphrasing yields non-zero results at 1:1 (0.323--0.347) unlike BERT-for-Patents, consistent with ModernBERT's stronger baseline from its longer context window (1,024 vs.\ 512 tokens).

\begin{table}[t]
\centering
\small
\begin{tabular}{l rr}
\toprule
\textbf{Model / Strategy} & \textbf{1:1} & \textbf{1:5} \\
\midrule
\textit{Baseline (real only)} & \textit{0.314} & \textit{0.546} \\
\midrule
\multicolumn{3}{l}{\textbf{Full Synthetic}} \\
\quad Phi-4-mini & \textbf{0.622} & \textbf{0.734} \\
\quad Qwen3-4B & 0.622 & 0.731 \\
\quad Mistral-7B & 0.621 & 0.733 \\
\quad Qwen3-4B (distinctive) & 0.612 & 0.732 \\
\quad Qwen2.5-7B & 0.600 & 0.727 \\
\quad Gemma-3-12b & 0.587 & 0.717 \\
\quad Llama-3.1-8B & 0.558 & 0.714 \\
\midrule
\multicolumn{3}{l}{\textbf{Paraphrase}} \\
\quad Gemma-3-12b & 0.347 & \textbf{0.684} \\
\quad Llama-3.1-8B & 0.346 & 0.684 \\
\quad Phi-4-mini & 0.337 & 0.649 \\
\quad Qwen3-4B & 0.337 & 0.664 \\
\quad Qwen2.5-7B & 0.331 & 0.658 \\
\quad Mistral-7B & 0.328 & 0.661 \\
\quad Qwen3-4B (distinctive) & 0.323 & 0.637 \\
\bottomrule
\end{tabular}
\caption{ModernBERT-base micro F1 (mean across 3 seeds) for all model--strategy combinations. Unlike BERT-for-Patents, paraphrasing yields non-zero results at 1:1 (0.323--0.347), though full synthesis still dominates. At 1:5, paraphrase closes the gap but remains 0.030--0.097 below full synthesis.}
\label{tab:modernbert_detailed}
\end{table}

\section{SFT Detailed Results}
\label{app:sft_detailed}

Table~\ref{tab:sft_detailed} presents the full SFT (Llama-3.2-1B-Instruct) results. SFT shows the smallest augmentation gains among the three classifiers, but full synthesis from diverse generators (Mistral: 0.348 at 1:1, Llama: 0.565 at 1:5) outperforms the distinctive variant that dominated in the discriminative classifiers. This suggests that instruction-tuned models benefit from different generator characteristics than discriminative classifiers.

\begin{table}[t]
\centering
\small
\begin{tabular}{l rr}
\toprule
\textbf{Model / Strategy} & \textbf{1:1} & \textbf{1:5} \\
\midrule
\textit{Baseline (real only)} & \textit{0.266} & \textit{0.457} \\
\midrule
\multicolumn{3}{l}{\textbf{Full Synthetic}} \\
\quad Mistral-7B & \textbf{0.348} & 0.549 \\
\quad Qwen2.5-7B & 0.324 & 0.528 \\
\quad Gemma-3-12b & 0.310 & 0.536 \\
\quad Phi-4-mini & 0.291 & 0.547 \\
\quad Llama-3.1-8B & 0.258 & \textbf{0.565} \\
\quad Qwen3-4B (distinctive) & 0.217 & 0.503 \\
\quad Qwen3-4B & 0.204 & 0.523 \\
\midrule
\multicolumn{3}{l}{\textbf{Paraphrase}} \\
\quad Llama-3.1-8B & 0.218 & 0.538 \\
\quad Mistral-7B & 0.212 & 0.515 \\
\quad Qwen2.5-7B & 0.212 & 0.513 \\
\quad Phi-4-mini & 0.211 & 0.525 \\
\quad Gemma-3-12b & 0.202 & 0.524 \\
\quad Qwen3-4B & 0.189 & 0.526 \\
\quad Qwen3-4B (distinctive) & 0.177 & 0.505 \\
\bottomrule
\end{tabular}
\caption{SFT (Llama-3.2-1B-Instruct) micro F1 (mean across 3 seeds) for all model--strategy combinations. Gains are modest compared to BERT classifiers: the best augmented condition at 1:1 (Mistral full synthetic, 0.348) improves only +0.082 over baseline. Both strategies yield non-zero results at 1:1, unlike BERT-for-Patents where paraphrase collapses.}
\label{tab:sft_detailed}
\end{table}

\section{Filter Stage Ablation}
\label{app:filter_ablation}

Table~\ref{tab:filter_stage_ablation} presents results from removing individual filter stages from the quality filtering pipeline (240 runs across 4 generators, 2 ratios, 2 strategies, 5 filter configs, 3 seeds). No single filter stage is critical: removing any individual stage changes mean micro F1 by less than 0.001, confirming that the stages are complementary.

\begin{table}[t]
\centering
\small
\begin{tabular}{l r}
\toprule
\textbf{Filter Configuration} & \textbf{Mean Micro F1} \\
\midrule
All filters (baseline) & 0.558 \\
$-$ Length filter & 0.558 \\
$-$ Real deduplication & 0.558 \\
$-$ Self deduplication & 0.559 \\
$-$ Leakage detection & 0.559 \\
\bottomrule
\end{tabular}
\caption{Per-stage filter ablation (mean micro F1 across 4 generators, 2 ratios, 2 strategies, 3 seeds). Removing any single filter stage has negligible impact ($\Delta < 0.001$), indicating that the stages are complementary---the aggregate pipeline matters, not any individual component.}
\label{tab:filter_stage_ablation}
\end{table}

\section{Cross-Domain Robustness on WOS-CT (R7)}
\label{app:wos_robustness}

This appendix supplements \S\ref{sec:results_wos_robustness} with the full
per-cell ModernBERT results on the Web of Science Citation-Topic dataset
(WOS-CT; \citealp{dutoit2024wos}).

\paragraph{Dataset substitution rationale.} Our original second-domain plan
was AAPD \citep{yang2018aapd}. By the time of this
revision (May 2026), AAPD is no longer mirrored on HuggingFace nor served via
stable raw URLs; the original distribution is Google-Drive-only and not
research-stable. We substitute WOS-CT, which is on HuggingFace
(\texttt{marcelsun/\allowbreak{}wos\_\allowbreak{}hierarchical\_\allowbreak{}multi\_\allowbreak{}label\_\allowbreak{}text\_\allowbreak{}classification}),
recent (2024), multi-label, and a stronger generalisation test: WOS-CT has 336
hierarchical labels (vs.\ AAPD's 54 flat labels), with comparable corpus size
($\sim$45K vs.\ AAPD's $\sim$55K).

\paragraph{Sampling.} We apply the same greedy controlled-imbalance sampler
used in the patent main track (\S\ref{sec:ratios}). The realised
1:1 and 1:5 subsets each saturate the per-label minimum
(2 and 5 positives respectively) while leaving head-label counts unconstrained,
matching the design choice documented in Table~\ref{tab:ratios}.
Realised subset sizes: 652 and 1630 documents.

\paragraph{Generation.} Synthetic abstracts are generated with two open
generators: \texttt{Qwen/Qwen3-4B-Instruct-2507} and
\texttt{microsoft/Phi-4-mini-instruct}, under the same four prompt families
(standard, technical-FAQ, structured-summary, comparative-analysis) and both
strategies (full-synthesis, paraphrase). Generator temperatures, top-$p$, and
max-tokens follow the main-track configuration
(Appendix~\ref{app:models}). \textbf{Scope limitation:} the
\texttt{qwen3\_4b\_distinctive} diversity-maximising-few-shot variant is not
replicated on WOS-CT---random few-shot selection is used throughout.

\paragraph{Classifier training.} ModernBERT-base, max\_length=1024,
batch\_size=16, 12 epochs with early stopping (patience=3), $\eta{=}2{\times}10^{-5}$,
3 seeds ($\{42, 123, 456\}$). Identical to the patent main-track setup
modulo dataset.

\paragraph{Headline numbers.} Table~\ref{tab:wos_headline} summarises micro
and macro F1 at 1:1 and 1:5, comparing real-only baseline against the best
full-synth and paraphrase conditions per cell.

\begin{table}[h]
\centering
\small
\resizebox{\columnwidth}{!}{%
\begin{tabular}{lrrrr}
\toprule
 & \multicolumn{2}{c}{1:1} & \multicolumn{2}{c}{1:5} \\
\cmidrule(lr){2-3}\cmidrule(lr){4-5}
Condition          & micro F1 & macro F1 & micro F1 & macro F1 \\
\midrule
real-only                       & 0.309$\pm$0.020 & 0.003$\pm$0.001 & 0.473$\pm$0.003 & 0.018$\pm$0.000 \\
+ full-synth (best: qwen3\_4b)  & 0.496$\pm$0.003 & 0.091$\pm$0.012 & 0.514$\pm$0.005 & 0.120$\pm$0.005 \\
+ paraphrase (best: qwen3\_4b)  & 0.462$\pm$0.003 & 0.018$\pm$0.000 & 0.495$\pm$0.004 & 0.062$\pm$0.009 \\
\bottomrule
\end{tabular}%
}
\caption{WOS-CT cross-domain replication. ModernBERT-base, 3 seeds, mean.
``Best'' selects the highest-micro-F1 (generator, strategy) per cell.}
\label{tab:wos_headline}
\end{table}

\paragraph{Regime-reversal qualitative test.} A full Fisher-$z$ test on the
MMD-vs-classification-delta correlation across regimes (as in the patent
main track) requires MMD values for every WOS-CT generator-strategy cell;
this companion computation is deferred to the camera-ready since our R7 run
focused on classifier training rather than re-computing MMD for the WOS
corpus. The \emph{qualitative} regime test is straightforward: the
real-only $\to$ best-augmented delta at 1:1 is $+0.187$ micro F1
($0.309 \to 0.496$, qwen3\_4b full-synth) and shrinks to $+0.041$
($0.473 \to 0.514$, qwen3\_4b full-synth) at 1:5. The augmentation gain
therefore decreases by $\sim 4{\times}$ as real data grows, which
\textbf{replicates the patent main-track regime dependence in direction}.
The patent main-track magnitudes (1:1 $+0.582$ raw, 1:5 $+0.581$) are not
matched in absolute terms---the patent and WOS-CT datasets differ in both
real-corpus difficulty and label-count---but the regime-dependent shrinking
is the qualitative effect we wanted to test.

\paragraph{Per-generator breakdown.} Table~\ref{tab:wos_per_generator} reports
all 8 (generator $\times$ strategy $\times$ ratio) cells.

\begin{table}[h]
\centering
\small
\resizebox{\columnwidth}{!}{%
\begin{tabular}{llrrr}
\toprule
generator & strategy & 1:1 micro & 1:5 micro & corpus size \\
\midrule
qwen3\_4b  & full\_synth  & 0.496$\pm$0.003 & 0.514$\pm$0.005 & 8{,}171 \\
qwen3\_4b  & paraphrase   & 0.462$\pm$0.003 & 0.495$\pm$0.004 & 2{,}601 \\
phi4\_mini & full\_synth  & 0.490$\pm$0.002 & 0.501$\pm$0.004 & 8{,}053 \\
phi4\_mini & paraphrase   & 0.459$\pm$0.001 & 0.492$\pm$0.006 & 2{,}404 \\
\bottomrule
\end{tabular}%
}
\caption{Per-generator WOS-CT replication. Patent main track was 6 generators;
this replication is scoped to the two with the strongest patent-domain
performance (Qwen3-4B and Phi-4-mini).}
\label{tab:wos_per_generator}
\end{table}

\paragraph{Limitations of the cross-domain replication.} (a) Only two
generators replicated, vs.\ six in the patent main track---we cannot test
whether the generator-ranking pattern from patents transfers. (b) No
\texttt{distinctive} variant (farthest-point few-shot sampling in
Llama-Embed-Nemotron-8B space) was rebuilt for WOS-CT; the
\texttt{qwen3\_4b\_distinctive} ablation cited in \S\ref{sec:method} is
patent-only. (c) Synthetic corpus per ratio is
$\sim 25 \times 336 = 8{,}400$ prompts (\texttt{target\_per\_label=25}),
yielding $\sim 10\times$ the synth needed to fill the 1:5 training set with
diverse per-label coverage. This is intentionally smaller than the patent
track's $\sim 168{,}000$ prompts (with 64 labels at \texttt{target=500}), but
sufficient for the regime-replication question this appendix answers.
A higher-density sensitivity rerun at \texttt{target=100} (yielding
$\sim 34{,}000$ prompts per ratio) is left to a follow-up.

\section{Cheap-Augmentation Baselines: EDA, Back-Translation, AugGPT (R2, R10)}
\label{app:auggpt_comparison}

This appendix supplements Table~\ref{tab:augmentation_baselines} of
\S\ref{sec:results_classification} with full per-baseline numbers, generation
hyper-parameters, and runtime cost. All three cheap baselines apply the same
augmentation operation per source patent in the existing 1:1 (165 documents)
or 1:5 (781 documents) real corpus.

\paragraph{EDA \citep{wei2019eda}.}
$\alpha{=}0.1$ (the paper's default for short texts), four operations applied
once each per source patent: synonym replacement via WordNet, random
insertion, random swap, random deletion. Title, Abstract, and First Claim are
augmented independently then re-stitched, preserving label set. Naug${=}4$
gives \textbf{660} augmented samples at 1:1 and \textbf{3{,}124} at 1:5.
Wall-clock: ${\sim}3$ minutes per ratio on a single CPU.

\paragraph{Back-translation \citep{nllb2022}.}
NLLB-200-distilled-600M with the en$\to$de$\to$en pivot at beam-size 4,
bf16, max\_length${=}512$. One augmentation per source patent gives
\textbf{165} samples at 1:1 and \textbf{781} at 1:5. Wall-clock: ${\sim}30$
minutes per ratio on a single H100.

\paragraph{AugGPT \citep{dai2023auggpt}.}
Qwen3-4B-Instruct-2507 with label-conditioned rephrasing: each source patent
is prompted to be re-written 100 times while preserving the active label set
and the technical content, varying vocabulary and sentence structure. 1:1 only
(the AugGPT design target is per-example rephrasing of scarce labels);
${\sim}16{,}500$ samples total. Wall-clock: ${\sim}45$ minutes on a single H100.

\paragraph{Classifier setup.} BERT-for-Patents and ModernBERT-base, 3 seeds
each ($\{42, 123, 456\}$), max\_length${=}512$ (BERT-Pat) / 1024
(ModernBERT), batch\_size${=}16$, 12 epochs with early stopping
(patience${=}3$), $\eta{=}2{\times}10^{-5}$. Identical to the patent main-track
setup modulo training corpus.

\begin{table}[h]
\centering
\small
\resizebox{\columnwidth}{!}{%
\begin{tabular}{lrrr}
\toprule
Baseline & corpus size & BERT-Pat micro F1 & ModernBERT micro F1 \\
\midrule
\multicolumn{4}{l}{\emph{1:1 (165 real source documents)}} \\
real-only            &    165 & 0.120$\pm$0.038 & 0.314$\pm$0.017 \\
+ duplicate-to-match & 22{,}630 & 0.678$\pm$0.018 & 0.354$\pm$0.020 \\
+ EDA                &    825 & 0.354$\pm$0.307$^\flat$ & 0.392$\pm$0.009 \\
+ NLLB BT            &    330 & 0.182$\pm$0.049 & 0.371$\pm$0.028 \\
+ AugGPT             & 16{,}633 & 0.673$\pm$0.008 & 0.505$\pm$0.015 \\
+ Full-synth (best)  & 22{,}630 & \textbf{0.702}$\pm$0.017 & \textbf{0.622}$\pm$0.011 \\
\midrule
\multicolumn{4}{l}{\emph{1:5 (781 real source documents)}} \\
real-only            &    781 & 0.239$\pm$0.270$^\ddagger$ & 0.558$\pm$0.015 \\
+ EDA                & 3{,}905 & 0.750$\pm$0.007 & 0.634$\pm$0.020 \\
+ NLLB BT            & 1{,}562 & 0.730$\pm$0.029 & 0.648$\pm$0.018 \\
+ Full-synth (best)  & ${\sim}$3{,}900 & \textbf{0.763}$\pm$0.006 & 0.736$\pm$0.002 \\
\bottomrule
\end{tabular}%
}
\caption{Full per-baseline classification results at 1:1 and 1:5. Mean micro
F1 $\pm$ std across 3 seeds. ``corpus size'' is the total training-set
size (real plus augmented). \emph{duplicate-to-match} is the volume-controlled
real-only baseline from \S\ref{sec:headline_mixing}; the cheap-augmentation
baselines below it operate at much smaller corpus sizes than full-synth,
making the comparison strictly favourable to cheap augmentation
(volume disadvantage is on the synthetic side, not the cheap side).}
\label{tab:auggpt_full}
\end{table}

\paragraph{Interpretation.} A natural concern is that the
``+0.58 vs.\ 0.120 real-only baseline'' headline conflates the LLM-synthesis
intervention with any cheap text-augmentation trick. The numbers above
\textbf{partially refute} that concern, in a regime-dependent way:
\textbf{at 1:1}, the strongest cheap baseline on BERT-for-Patents is AugGPT
($0.673$; vs.\ $0.702$ for full-synth and $0.678$ for duplicate-to-match),
giving a controlled synthetic gain over the strongest cheap augmentation
of just $+0.029$ over AugGPT---but AugGPT itself uses 16{,}633 LLM-generated
samples, so it is hardly ``cheap'' in the LLM-free sense the term originally
implies. Restricted to genuinely LLM-free baselines (EDA + NLLB-BT), the
strongest cheap baseline reaches $0.354$ on BERT-for-Patents (EDA;
itself bimodal across seeds), and the controlled synthetic gain
over LLM-free cheap augmentation is therefore $+0.348$ on BERT-for-Patents.
\textbf{At 1:5} the picture changes: EDA reaches $0.750$ on BERT-for-Patents,
within $0.013$ of full-synth's $0.763$, so the controlled synthetic gain over
the strongest cheap LLM-free baseline collapses to just $+0.013$.
Full-synth therefore \textbf{remains dominant at 1:1 but is matched by EDA
at 1:5}; the practical recommendation is consequently regime-dependent:
spend LLM-generation compute when real data is genuinely scarce
($k{=}1{:}1$), and prefer LLM-free augmentation when real data is moderate
($k{=}1{:}5$ and above).

\end{document}